\documentclass{article} 
\usepackage{iclr2026_conference,times}

\usepackage{amsmath,amsfonts,bm}

\def\eqref#1{equation~\ref{#1}}

\def\1{\bm{1}}

\DeclareMathAlphabet{\mathsfit}{\encodingdefault}{\sfdefault}{m}{sl}
\SetMathAlphabet{\mathsfit}{bold}{\encodingdefault}{\sfdefault}{bx}{n}

\iclrfinalcopy
\usepackage{hyperref}
\usepackage{url}

\usepackage{siunitx}
\usepackage[table]{xcolor}
\usepackage{caption}
\usepackage{multirow}
\usepackage{graphicx}
\usepackage{url}            
\usepackage{booktabs}       
\usepackage{amsfonts}       
\usepackage{nicefrac}       
\usepackage{microtype}      
\usepackage[dvipsnames]{xcolor}
\usepackage{amsmath, amssymb, amsthm}
\usepackage{booktabs}
\usepackage{siunitx}
\usepackage{graphicx}
\usepackage{xcolor}
\usepackage{multicol}
\usepackage{colortbl}
\usepackage{tabularx}
\usepackage{makecell}
\usepackage{multirow}
\usepackage{diagbox}
\usepackage{enumitem}
\usepackage{tgcursor}
\usepackage{svg} 
\usepackage{placeins}
\usepackage{algorithm}
\usepackage{algorithmic}
\usepackage{newfloat}
\usepackage{listings}
\usepackage{wrapfig}
\usepackage{marvosym}

\DeclareCaptionStyle{ruled}{labelfont=normalfont,labelsep=colon,strut=off} 
\floatstyle{ruled}
\newfloat{listing}{tb}{lst}{}
\floatname{listing}{Listing}
\newcommand{\mymethod}{EssenceBench}

\newtheorem{definition}{Definition}
\definecolor{inc}{HTML}{1F77B4}   
\definecolor{dec}{HTML}{D62728}   
\definecolor{metaBG}{HTML}{E6FFCC}   
\definecolor{essBG}{HTML}{CCF2FF}    
\definecolor{authorcolor}{HTML}{e64159}
\newcommand{\valmeta}[1]{\makebox[14mm][r]{#1\phantom{$\uparrow$}}}

\title{Rethinking LLM Evaluation: Can We Evaluate LLMs with 200$\times$ Less Data?}

\author{
    \small
  {\bf
    \hspace{-10pt}Shaobo Wang $^{\dagger}$$^*$$^{{\color{authorcolor}\boldsymbol{1,2,3}}}$
    Cong Wang
    $^*$$^{{\color{authorcolor}\boldsymbol{1}}}$
    Wenjie Fu
    $^*$$^{{\color{authorcolor}\boldsymbol{1,4}}}$
    Yue Min
    $^{{\color{authorcolor}\boldsymbol{1,5}}}$    
    Mingquan Feng
    $^{{\color{authorcolor}\boldsymbol{2}}}$
    Isabel Guan
    $^{{\color{authorcolor}\boldsymbol{5}}}$
    Xuming Hu
    $^{{\color{authorcolor}\boldsymbol{5,6}}}$
  } \\
    \small
    {
    \bf
    \hspace{-10pt}Conghui He
    $^{{\color{authorcolor}\boldsymbol{7}}}$
    Cunxiang Wang
    $^{{\color{authorcolor}\boldsymbol{8}}}$
    Kexin Yang
    $^{{\color{authorcolor}\boldsymbol{3}}}$
    Xingzhang Ren
    $^{{\color{authorcolor}\boldsymbol{3}}}$
    Fei Huang
    $^{{\color{authorcolor}\boldsymbol{3}}}$
    Dayiheng Liu
    $^{{\color{authorcolor}\boldsymbol{3}}}$
    Linfeng Zhang
    $^{\text{\Letter}}$$^{{\color{authorcolor}\boldsymbol{1,2}}}$
    \vspace{2pt}
  } \\
    \small
    {
    $^{\color{authorcolor}\boldsymbol{1}}$ EPIC Lab, SJTU $\quad$
    $^{\color{authorcolor}\boldsymbol{2}}$ SJTU \ \
    $^{\color{authorcolor}\boldsymbol{3}}$ Alibaba Group \ \ 
    $^{\color{authorcolor}\boldsymbol{4}}$ FDU \ \
    $^{\color{authorcolor}\boldsymbol{5}}$ HKUST \ \
    $^{\color{authorcolor}\boldsymbol{6}}$ HKUST (GZ)
    }\\
    \small
    {
    $^{\color{authorcolor}\boldsymbol{7}}$ Shanghai AI Lab \ \
    $^{\color{authorcolor}\boldsymbol{8}}$ ZhipuAI \ \
    * Equal contribution \ \
    \Letter\ Corresponding author  \ \
    $\dagger$\ Project Head
    }
}

\begin{document}

\maketitle

\begin{abstract}
As the demand for comprehensive evaluations of diverse model capabilities steadily increases, benchmark suites have correspondingly grown significantly in scale. Despite notable advances in redundancy reduction and subset-level performance prediction, a systematic framework that effectively integrates these methods to ensure both prediction accuracy and ranking consistency is still largely elusive. In this paper, we first perform a sample-level analysis of benchmark redundancy and identify several highly similar samples that can be eliminated. Besides, we frame benchmark compression as an optimization problem with the aim of score reconstruction. Building on these, we then propose \mymethod{}, a coarse-to-fine framework utilizing an iterative Genetic Algorithm (GA), which takes the advantages of fitness-based subset search and attribution-based sample search. Compared to previous methods, our approach yields superior compression results with lower reconstruction error and markedly higher efficiency. In particular, on the HellaSwag benchmark (10K samples), our method preserves the ranking of all models shifting within 5\% using 25$\times$ fewer samples, and achieves 95\% ranking preservation shifting within 5\% using only 200$\times$ fewer samples. 
\end{abstract}
\begin{figure}[h] 
    \centering
    \vspace{-15pt}
    \rotatebox{0}{%
        \includegraphics[width=0.99\linewidth]{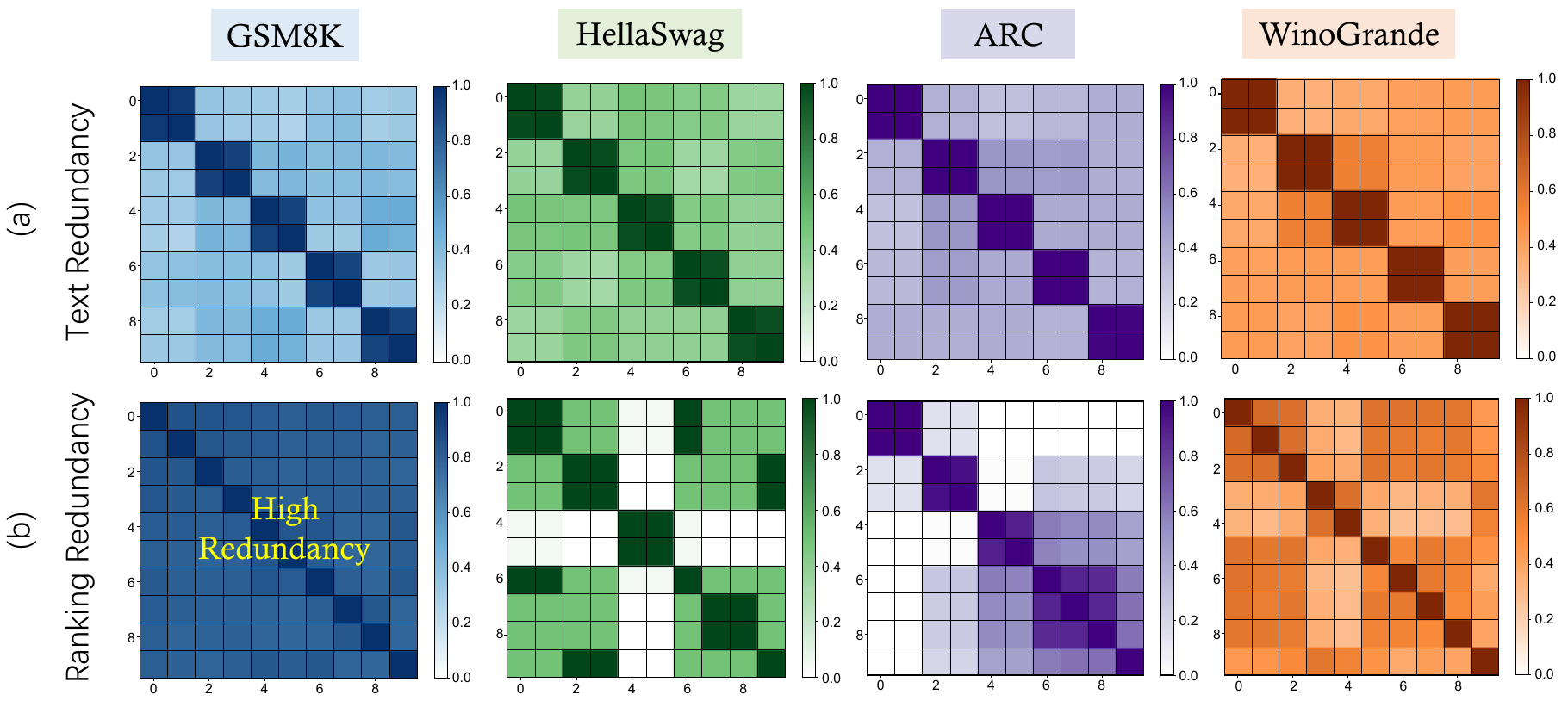} 
    }
    \vspace{-10pt}
    \caption{Prevalent redundancy across widely used benchmark datasets. Based on 10 randomly sampled instances per dataset, panel (a) depicts the \textit{text embedding similarity} (Definition~\ref{def:text_redundancy}) , reflecting semantic overlap among instances, and panel (b) presents the \textit{ranking embedding similarity} (Definition~\ref{def:ranking_redundancy}), measured through consistency of model performance rankings across sampled subsets.
    }
    \label{fig:redundancy_pattern}
    \vspace{-5pt}
\end{figure}
\section{Introduction}
\label{sec:intro}
In recent years, large language models (LLMs) have advanced rapidly with the release of models such as GPT-4~\citep{GPT4}. This swift progress has led to a shift in increasingly sophisticated LLM benchmark design, moving from traditional natural language processing (NLP) tasks to more comprehensive, multidimensional evaluation suites. Prominent benchmarks have been released to evaluate LLMs in areas such as multilingual understanding~\citep{zhaolarge}, long-context reasoning~\citep{kuratov2024babilong}, instruction following~\citep{yin2023llm}, mathematical reasoning~\citep{shao2024deepseekmath}, code comprehension and generation~\citep{nam2024using}, multidisciplinary knowledge acquisition~\citep{zhang2025redundancy}, and tool integration~\citep{chen2024advancing}.
However, as the scope and granularity of evaluation expand, so does the scale and computing cost of the benchmarking process. For instance, \texttt{OpenCompass}~\citep{Opencompass} integrates over 60 subtasks across more than 25 capability dimensions. {Evaluating Qwen2.5-7B-Instruct\footnote{https://huggingface.co/Qwen/Qwen2.5-7B-Instruct} across all these tasks often takes about \textbf{1k GPU hours}, consuming millions to tens of millions of tokens. Consequently, the question of \textbf{{how to efficiently reduce the sample size of benchmark datasets while preserving the reliability of evaluation}} has become a critical challenge in the current phase of LLM.

\begin{figure}[tb!] 
    \centering
    \includegraphics[width=1.0\linewidth]{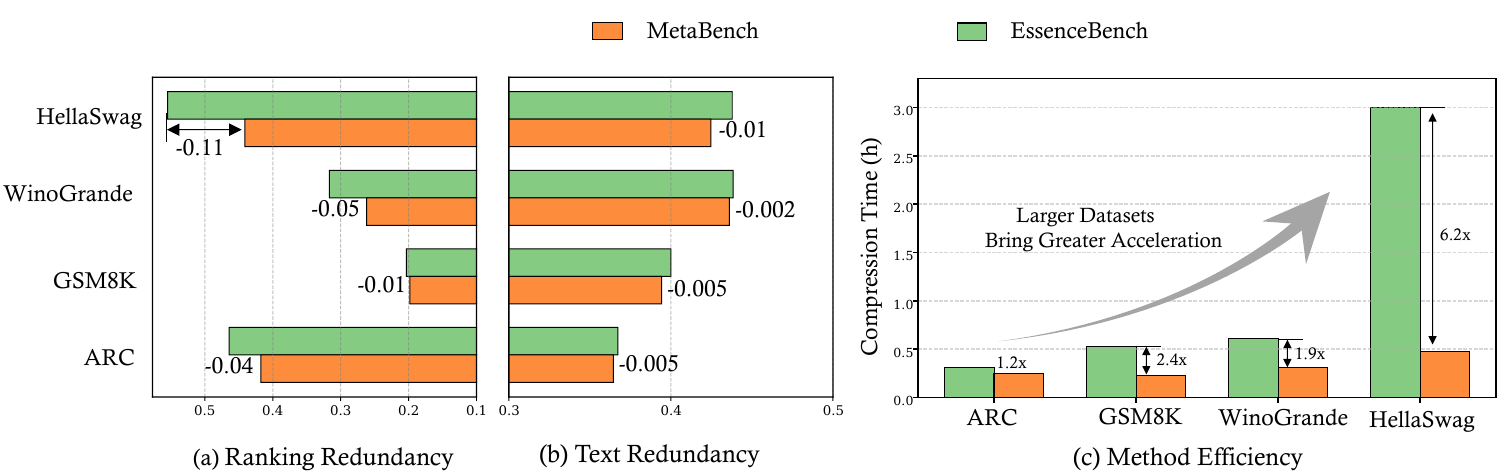} 
    \caption{Comparison of existing benchmark compression approaches and our \mymethod{}. (a) ranking and (b) text redundancy comparison and (c) compression time comparison.}
    \label{fig:redundancy_efficiency}
    \vspace{-15pt}
\end{figure}

We first revisit the foundations of LLM benchmark evaluation, focusing on a critical yet under-examined phenomenon: \textit{sample redundancy} in the \texttt{Open LLM Leaderboard}\footnote{https://huggingface.co/datasets/open-llm-leaderboard-old/results}~\citep{OpenLLMLeaderBoard}. Our analysis quantifies redundancy through two complementary dimensions: (i) \textbf{Text-level redundancy}: Defined as lexical and semantic overlap between evaluation instances (Definition~\ref{def:text_redundancy})
and (ii) \textbf{Ranking-level redundancy}: Measured through consistency of model performance rankings across sampled subsets (Definition~\ref{def:ranking_redundancy}). As shown in Figure~\ref{fig:redundancy_pattern}, we systematically evaluate redundancy across benchmark datasets by randomly sampling 10 instances per dataset. Our analysis reveals significant redundancy patterns that persist across diverse benchmark configurations and model architectures, manifesting in both textual content and performance ranking dimensions.

Recent studies have explored benchmark compression. Methods such as LLM-based annotation~\citep{buildbench}, active sample querying~\citep{Kossen_Farquhar_Gal_Rainforth_2021, tinybench}, and psychological approaches~\citep{tinybench, MetaBench} have been employed to reduce benchmark dataset size while maintaining evaluation quality. Notably, MetaBench~\citep{MetaBench} uses Generalized Additive Models (GAM)~\citep{GAM} to model the relationship between subset scores and full set performance, and it employs root mean square error (RMSE) as an index to guide sampling policies. Despite their contributions, previous works have two significant limitations: 
\begin{enumerate}[leftmargin=*, topsep=2pt, itemsep=2pt, parsep=0pt]
   \item \textbf{Neglect of Sample Interactions}: Conventional approaches treat test samples as independent entities, ignoring semantic relationships. Specifically, when two samples exhibit high similarity, their evaluation outcomes often correlate strongly, suggesting a redundancy that warrants systematic elimination. Developing robust metrics and principled approaches for identifying such redundancies remains an open challenge.
   
   \item \textbf{Inefficient Search Mechanisms}: Existing compression techniques rely on statistical or heuristic methods (\emph{e.g.}, GAM, LLM scoring, active querying) that suffer from high computational overhead or suboptimal convergence.
\end{enumerate}

\begin{figure*}[tb!]
    \centering
    \vspace{-5pt}
    \includegraphics[width=0.9\linewidth]{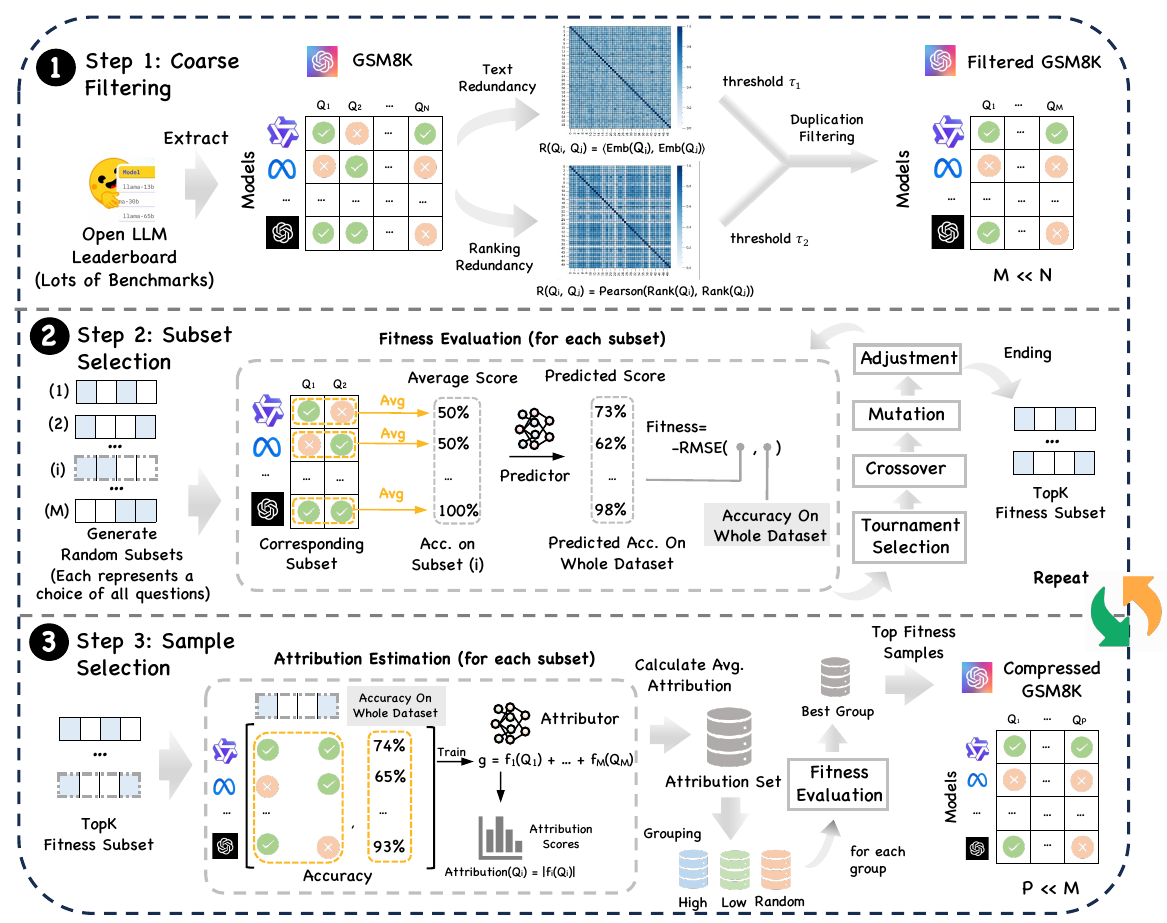}
    \vspace{-6pt}
    \caption{The pipeline of \mymethod{}. (I) \textbf{Coarse Filtering}. By extracting the binary score matrix for each benchmark and computing both text-level and ranking-level redundancies,  samples that exceed thresholds are removed. (II) \textbf{Subset Selection}. Genetic Algorithm (GA) is applied beginning with generating random subsets. With fitness evaluated by the error of predicted accuracy, subsets are optimized via fitness-based tournament selection, crossover, mutation, and adjustment. (III) \textbf{Sample Selection}. Attribution of each sample is estimated from the top-performing subsets by utilizing weights when training a model. According to that, samples are divided into groups. GA is then reapplied within each group to identify the most representative and informative subset.}
    \label{fig:pipeline}
    \vspace{-10pt}
\end{figure*}

To tackle the challenge of redundant and costly LLM benchmark evaluations, as shown in Figure~\ref{fig:redundancy_efficiency}, we introduce \mymethod{}, \textit{a coarse-to-fine, iterative compression framework} that preserves evaluation fidelity while reducing dataset size. As illustrated in Figure~\ref{fig:pipeline}, \textbf{(i)} we first extract each benchmark’s score matrix from the \texttt{Open LLM Leaderboard}. To weight the interactions between samples and eliminate the first limitation, we quantify text-level and ranking-level redundancies via embedding similarities and ranking correlations, then eliminate duplicate samples. \textbf{(ii)} On this filtered set, we launch an iterative Genetic Algorithm (GA) to identify compact yet representative subsets: in each GA run, candidate subsets are evaluated by a predictor model to minimize the prediction error against full-dataset accuracy, guiding the search toward subsets that faithfully reconstruct overall performance. \textbf{(iii)} Finally, to resolve the inefficiency of search mechanisms, we refine our selection through sample-level attribution, using the weights learned during model training to partition instances into high, low, and random attribution groups; we reapply GA within each group and choose the top-performing samples from the best group as our compressed benchmark for that round. This further searching step introduces sample-level diversity thus alleviating suboptimal convergence. By combining redundancy-aware filtering with iterative GA optimization, \mymethod{} delivers benchmarks that are both lean and reliable, enabling rapid, cost-effective evaluation of cutting-edge LLMs. Our contributions are as follows:
\begin{itemize}[leftmargin=10pt, topsep=0pt, itemsep=1pt, partopsep=1pt, parsep=1pt] 
    \item We systematically analyze  redundancy problems in LLM benchmarks and observe that all benchmarks in \texttt{Open LLM Leaderboard} share a sample redundancy phenomenon, which causes evaluation inefficiency.
    \item We frame benchmark compression as an optimization problem and tackle with it effectively by combining redundancy-based coarse filtering and iterative Genetic Algorithm. The proposed framework, \mymethod{}, efficiently addresses this problem while ensuring scalability.
    \item Experimental results demonstrate that \mymethod{} achieves significant reductions while efficiently maintaining rankings. Notably, on HellaSwag (10K samples), our method preserves the ranking of all models shifting within 5\% using 25$\times$ fewer samples.

\end{itemize}

\section{Related Works}
\noindent \textbf{Large Language Model Evaluation.} Benchmarks are indispensable for measuring LLM capabilities and catalyzing research. Standard NLU and reading comprehension tasks include \texttt{GLUE}~\citep{GLUE} and \texttt{SQuAD}~\citep{SQuAD}; mathematical reasoning is tested by \texttt{GSM8K}~\citep{GSM8K}, \texttt{MATH}~\citep{MATH} and \texttt{Mathqa}~\citep{Mathqa}; coding ability via \texttt{HumanEval}~\citep{humaneval} and \texttt{MBPP}~\citep{MBPP}; instruction following by \texttt{IFEval}~\citep{IFEval}; and multiple‐choice and commonsense reasoning by \texttt{MMLU}~\citep{MMLU}, \texttt{ARC-Challenge}~\citep{ARC} and \texttt{HellaSwag}~\citep{HelloSwag}. Platforms such as \texttt{Open LLM Leaderboard}~\citep{OpenLLMLeaderBoard} and \texttt{OpenCompass}~\citep{Opencompass} provide unified pipelines and live leaderboards, but rarely consider the cost and efficiency.


\noindent \textbf{LLM Benchmark Compression.} As benchmarks scale up, compact evaluation suites are critical~\citep{buildbench}. TinyBenchmark~\citep{tinybench} combines statistical selection with Item Response Theory (IRT) to pick 100 representative items; MetaBench~\citep{MetaBench} uses IRT and Fisher information over 5000 LLM outputs to select discriminative examples, achieving an average RMSE of $1.5\%$. However, metaheuristic search techniques such as Genetic Algorithms have not been explored, and the redundancy phenomenon highlighted in~\citep{zhang2025redundancy, li2025information} remains insufficiently addressed. For a broader survey of recent advances in LLM data selection and compression, we refer readers to Appendix ~\ref{sec:more_related works}.

\section{Methodology}



\subsection{Preliminaries: Benchmark Compression}
\label{subsubsec: def}

The benchmark compression problem can be framed as selecting a $k$-element subset (a coreset) from the universal set (the benchmark). If the goal is to reconstruct the score of the benchmark, finding the subset requires searching over an exponentially large, discrete space of candidate subsets, which makes it a classic NP-hard combinatorial optimization task. Formally, it can be defined as:
\begin{definition}[Benchmark Compression]
\label{def:benchmark compression}
Let $\mathcal{D} = \{x_1,\dots,x_N\}$ be a benchmark dataset, and let $g:2^{\mathcal{D}}\to \mathbb{R}$
be an aggregate scoring function that assigns a performance score to any subset of $\mathcal{D}$. Given a budget $k < N$, the compression problem seeks a $k$-element subset $\tilde{\mathcal{D}} \subseteq \mathcal{D}$ that best reconstructs the full-dataset score:
\begin{equation}
\setlength{\abovedisplayskip}{3pt}
\setlength{\belowdisplayskip}{3pt}
\tilde{\mathcal{D}}^* = \mathop{\arg\min}_{\tilde{\mathcal{D}} \subseteq \mathcal{D}, |\tilde{\mathcal{D}}| = k}
\mathcal{L}\bigl(g(\mathcal{D}),\, g(\tilde{\mathcal{D}})\bigr),
\end{equation}
where $\mathcal{L}(\cdot,\cdot)$ represents a suitable error measure.
\end{definition}
However, Definition~\ref{def:benchmark compression} relies on a function to impractically traverse all subsets of the whole dataset.
To effectively overcome this limitation, we leverage the fact that public leaderboards~\citep{OpenLLMLeaderBoard} report a binary correctness (right or wrong) for each model on each benchmark sample. By systematically organizing these outcomes into a \textit{score matrix}, we simplify the scoring function into column selection and accuracy computation. 
Let $\mathcal{D}$ be the benchmark and $N_\mathrm{LLM}$ denote the number of LLMs tested on it. The \emph{score matrix} is denoted as: $ \mathbf{S}\in\{0,1\}^{N_\mathrm{LLM} \times N}
$, where $\mathbf{S}_{i,j}$ explicitly denotes the score of $\mathrm{LLM}_i$ on sample $x_j$, 1 for right and 0 for wrong. Let $\mathbf{y} \in \mathbb{R}^{N_\mathrm{LLM}}$ denote the accuracy of LLMs on $\mathcal{D}$, where $y_i = \frac{1}{N}\sum_{j=1}^N\mathbf{S}_{i,j}$ is the accuracy of $\mathrm{LLM}_i$. The concrete formulation of benchmark compression is then defined in Definition~\ref{def:concrete formulation of benchmark compression}.
\begin{definition}[Concrete Formulation of Benchmark Compression]
\label{def:concrete formulation of benchmark compression}

Let a binary mask represents the selection of a $k$-element subset:
$
\mathbf{m} \in \{0,1\}^N\;\; \text{s.t.} \sum_{j=1}^N m_j = k
$.
By indexing the mask, the matching columns of $\,\mathbf{S}$ can be got, which is denoted as $\mathbf{S}_{\mathbf{m}}$. The aggregate scoring function $g$ in Definition~\ref{def:benchmark compression} can be concretized as: $g(\tilde{\mathcal{D}}) = g(\mathbf{S_{\mathbf{m}}})$. Therefore, the optimization problem is as:
\begin{equation}
\setlength{\abovedisplayskip}{3pt}
\setlength{\belowdisplayskip}{3pt}
\label{eq:compression}\min_{\mathbf{m} \in \{0,1\}^N} \;
\mathcal{L}\bigl(\mathbf{y},\, g(\mathbf{S}_{\mathbf{m}})\bigr) \;\;\;\;
\mathrm{s.t.}
\sum_{j=1}^N m_j = k.
\end{equation}
\end{definition}

\subsection{A Closer Look at the Sample Redundancy Phenomenon in LLM Benchmarks}
\label{subsec: red}
As illustrated in Figure~\ref{fig:redundancy_pattern}, many LLM benchmarks exhibit high overlap both in the text of their prompts and in the models’ performance rankings in our quantification of sample redundancy, \emph{i.e.} text redundancy (Definition~\ref{def:text_redundancy}) and ranking redundancy (Definition~\ref{def:ranking_redundancy}). Intuitively, when two examples share similar wording or contain nearly identical behavior across some models, retaining both adds little new information while doubling evaluation costs.
\begin{definition}[Sample Redundancy from Text Perspective]
\label{def:text_redundancy}

Let $\mathcal{D} = \{x_1, x_2, \ldots, x_N\}$ be a benchmark dataset where each $x_i$ denotes the textual input. Let $\mathrm{Emb}: x \mapsto \mathbb{R}^{d_\mathrm{emb}}$ denote an embedding mapping of the input. The redundancy of a specific sample pair is defined as:
\begin{equation}
\setlength{\abovedisplayskip}{3pt}
\setlength{\belowdisplayskip}{3pt}
\mathcal{R}_{\mathrm{text}}(i, j) = \langle \mathrm{Emb}(x_i), \mathrm{Emb}(x_j) \rangle.
\end{equation}
The redundancy of a sample is then defined as:
\begin{equation}
\setlength{\abovedisplayskip}{3pt}
\setlength{\belowdisplayskip}{3pt}
\mathcal{R}_{\mathrm{text}}(i) = \frac{\sum_{j\ne i}\langle \mathrm{Emb}(x_i) , \mathrm{Emb}(x_j)\rangle}{N-1}.
\end{equation}
The overall redundancy of the benchmark is defined as the average redundancy across all samples:
\begin{equation}
\setlength{\abovedisplayskip}{3pt}
\setlength{\belowdisplayskip}{3pt}
\mathcal{R}_{\mathrm{text}}(\mathcal{D}) = \frac{\sum_{i=1}^N \mathcal{R}_{\mathrm{text}}(i)}{N}.
\end{equation}
\end{definition}

\begin{definition}[Sample Redundancy from Ranking Perspective]
\label{def:ranking_redundancy}

Let $\mathcal{D} = \{x_1, x_2, \ldots, x_N\}$ be a benchmark dataset. Suppose we are given a set of responses from several LLMs, and for each sample $x_i$, we define a ranking score $r_i \in \mathbb{R}$ indicating the model's confidence or correctness on that sample. For example, $r_i$ could be a binary indicator (e.g., correct/incorrect) or a continuous score (e.g., answer log-likelihood). We define the redundancy between two samples $x_i$ and $x_j$ in terms of their ranking correlation as:
\begin{equation}
\setlength{\abovedisplayskip}{3pt}
\setlength{\belowdisplayskip}{3pt}
\mathcal{R}_{\mathrm{ranking}}(i, j) = \rho(r_i, r_j)
\end{equation}
where $\rho(\cdot, \cdot)$ denotes the correlation coefficient, which can be Pearson, Spearman or $R^2$. The one-sample and overall redundancy are defined the same way as:
\begin{equation}
\setlength{\abovedisplayskip}{3pt}
\setlength{\belowdisplayskip}{3pt}
\mathcal{R}_{\mathrm{ranking}}(i) = \frac{\sum_{j\ne i} |\rho(r_i, r_j)|}{N-1},\quad \mathcal{R}_{\mathrm{ranking}}(\mathcal{D}) = \frac{\sum_{i=1}^N |\rho(r_i, r_j)|}{N}.
\end{equation}
\end{definition}
These two definitions are intuitively derived from both human and LLM perspectives. Specifically, textual redundancy quantifies semantic similarity, while ranking redundancy assesses behavioral similarity. When combined, they reveal complementary overlaps that each alone would overlook, facilitating more systematic redundancy elimination and benchmark pruning.

\subsection{\mymethod{}}
\label{subsec: ga}

\noindent{\textbf{Step1: Coarse Filtering}}. On the basis of 
\textit{text redundancy} (Definition~\ref{def:text_redundancy}) and \textit{ranking redundancy} (Definition~\ref{def:ranking_redundancy}). 
The benchmark $\mathcal{D}$ with size $N$ can be filtered through the threshold $\tau_{\mathrm{text}}$ and $\tau_{\mathrm{ranking}}$ to size $M$ as shown in Figure~\ref{fig:pipeline}. We examine the dataset in its original order and decide for each sample $x_i$ whether to keep or discard it. For each sample $x_j$, if either $\mathcal{R}_{\mathrm{text}}(j,i) > \tau_{\mathrm{text}}$ or $ \mathcal{R}_{\mathrm{ranking}}(j,i) > \tau_{\mathrm{ranking}}$, then $x_i$ is discarded; otherwise, it is retained. This ensures that among any highly redundant pair, the sample encountered first is always kept. Let $\epsilon_i$ be the flag indicating whether $x_i$ should be discarded, formalized as:
\begin{equation}
\setlength{\abovedisplayskip}{3pt}
\setlength{\belowdisplayskip}{3pt}
\epsilon_i = \prod_{j=1}^{i-1} \mathbf{1}\!\bigl(\mathcal{R}_{\mathrm{text}}(j,i)\le\tau_{\mathrm{text}} \;\wedge\; \mathcal{R}_{\mathrm{ranking}}(j,i)\le\tau_{\mathrm{ranking}}\bigr)\,,
\end{equation}
where $\mathbf{1(\cdot)}$ denotes the indicator function. Therefore, the filtered benchmark is
$\mathcal{D}_\mathrm{filtered} \;=\; \{\,x_i \mid \epsilon_i = 1\}$ with size $M$. In this process, we strip away the most redundant examples to yield a compact yet representative filtered set, lightening our evaluation load and making compression faster.

\noindent{\textbf{Step2: Fitness-based Subset Selection}}.
\label{subsubsec: imp} 
To shrink the filtered benchmark while preserving its ability to approximate the benchmark score, we employ an iterative genetic algorithm~\citep{ga1, ga2, ga3, ga4} as a heuristic search over the space of possible subsets. As Figure~\ref{fig:pipeline} shows, starting from a randomly initialized \emph{population} of $k$-element masks, we repeatedly evaluate each mask’s \emph{fitness}, select parents via \emph{tournament selection}, generate offspring through \emph{crossover} and \emph{mutation}, and then \emph{adjust} each child to enforce the $k$-element constraint. Over successive generations, this process converges toward high-quality subsets that minimize reconstruction error. The whole process is illustrated in Algorithm~\ref{alg:ga_step2} in Appendix ~\ref{sec:Pseudo Code}. Detailed descriptions of each step are provided below.

\noindent \textbf{Individual and Population}. An \textit{individual} is a mask $\mathbf{m} \in \{0,1\}^M\; \text{s.t.}\sum_{j=1}^M m_j = k$, which is also called a $subset$. The \textit{population} is denoted as a group of individuals:
$\mathcal{P} = \{\mathbf{m}^{(1)}, \ldots, \mathbf{m}^{(N_\mathcal{P})}\}$, where $N_{\mathcal{P}}$ is the number of the individuals in the population. The final top-$N_{\mathcal{E}}$ best subsets are also denoted as a set: $\mathcal{E} = \{\mathbf{m}^{(1)}, \ldots, \mathbf{m}^{(N_{\mathcal{E}})}\}$.


\noindent \textbf{Fitness Evaluation}. The process of \textit{fitness evaluation} is the same as minimizing $\mathcal{L}$ in Equation~\ref{eq:compression}. To calculate, a general additive model (GAM)~\citep{GAM} is trained to act as the aggregate scoring function $g$ in Definition~\ref{def:concrete formulation of benchmark compression}. The training data $\mathcal{T}$ and validation data $\mathcal{V}$ are denoted as \(\{s_i,y_i\}_{i\in\mathcal{T}}\) and \(\{s_i,y_i\}_{i\in\mathcal{V}}\), which constructs a map from subset accuracy to whole dataset accuracy. Let the score matrix and the accuracy of LLMs on $\mathcal{D_{\mathrm{filtered}}}$ is denoted as: $\mathbf{S}_\mathrm{filtered} \in \{0,1\}^{N_\mathrm{LLM} \times M}, \mathbf{y_\mathrm{filtered}} \in \mathbb{R}^{N_\mathrm{LLM}}$. Therefore, $y_i$ can be directly obtained from $\mathbf{y}$, $s_i$ can be calculated as:
$s_i=\frac{1}{k}\sum_{i=1}^{M}\mathbf{S}_{\mathrm{filtered}}\mathbf{m}^{(i')}$. Since RMSE is used as the error measure, given an individual $\mathbf{m}$, its fitness is calculated as:
\begin{equation}
\setlength{\abovedisplayskip}{3pt}
\setlength{\belowdisplayskip}{3pt}
\label{eq:fitness}
\begin{aligned}
\mathrm{fitness}\bigl(\mathbf{m}\bigr)
& =-\mathrm{RMSE}\bigg(\mathbf{y},\, g(\mathbf{S}_{\mathbf{m}})\bigg)=\sqrt{\frac{1}{|\mathcal{V}|}\sum_{j\in\mathcal{V}}
\bigl(\hat y_j-y_j\bigr)^2}.
\end{aligned}
\end{equation}
where $\hat{y}_j = g(\mathbf{s}_j)$ is the GAM-predicted accuracy for the $j$-th individual based on its subset score $\mathbf{s}_j$.

\noindent \textbf{Tournament Selection}. In this process, we aim to choose an individual from $\mathcal{P}$ as a \textit{parent} according to fitness (Equation~\ref{eq:fitness}). Let 
\( \mathbf{m}^{(a)}, \mathbf{m}^{(b)} \) denote the two parents chosen in this process.

\noindent \textbf{Crossover}. In this process, the goal is to get a new individual by combining the information of the two parents. Let $\xi_j$ denote a flag which parent the new individual should follow, the \textit{crossover} process generates a new individual $\mathbf{m}^{(c)}$ by:
\begin{equation}
\setlength{\abovedisplayskip}{3pt}
\setlength{\belowdisplayskip}{3pt}
m^{(c)}_j = (m_j^{(a)} \land \xi_j) \lor (m_j^{(b)} \land \neg \xi_j),
\end{equation}
where $ \xi_j \sim \text{Bernoulli}(0.5), \; j \in [1, M]$.

\begin{table*}[tb!]
\vspace{-10pt}
\caption{Prediction Error ($\downarrow$) of selected subsets with different sizes.} 
\label{tab:rmse}
\vspace{-5pt}
\centering
\resizebox{.9\textwidth}{!}{
\begin{tabular}{@{}c|c|cccccccccc@{}}
\toprule
\multirow{2}{*}{Dataset} & \multicolumn{1}{c|}{\multirow{2}{*}{Method}} & \multicolumn{10}{c}{Coreset Size} \\ \cmidrule(l){3-12} 
 &  & 50 & 100 & 150 & 200 & 250 & 300 & 350 & 400 & 450 & 500 \\ \midrule
\multirow{5}{*}{GSM8K} & Random & 3.6894 & 2.8531 & 2.2934 & 1.9454 & 1.6267 & 1.4013 & 1.3432 & 1.2279 & 1.1487 & 1.0400 \\
 & PPL & 4.1941 & 2.6987 & 2.3007 & 1.9218 & 1.719 & 1.5153 & 1.3588 & 1.3002 & 1.1988 & 1.1301 \\
 & GraNd & 3.8516 & 2.8929 & 2.3725 & 2.0956 & 1.8256 & 1.5994 & 1.4442 & 1.2719 & 1.1568 & 1.0692 \\
 & MetaBench & 3.5283 & 2.4335 & 2.0673 & 1.7597 & 1.5529 & 1.3873 & 1.2631 & 1.1301 & 1.0333 & 0.9579 \\
 & \cellcolor{green!5}\mymethod{} & \cellcolor{green!5}\textbf{2.7685} & \cellcolor{green!5}\textbf{1.6671} & \cellcolor{green!5}\textbf{1.1516} & \cellcolor{green!5}\textbf{0.8635} & \cellcolor{green!5}\textbf{0.7181} & \cellcolor{green!5}\textbf{0.6101} & \cellcolor{green!5}\textbf{0.5588} & \cellcolor{green!5}\textbf{0.5037} & \cellcolor{green!5}\textbf{0.4561} & \cellcolor{green!5}\textbf{0.3769} \\ \midrule
\multirow{5}{*}{ARC} & Random & 3.1528 & 2.4463 & 2.1646 & 1.7307 & 1.4499 & 1.3194 & 1.1533 & 1.0976 & 0.9341 & 0.9015 \\
 & PPL & 5.3343 & 3.1191 & 2.1166 & 1.8187 & 1.5459 & 1.4156 & 1.2165 & 1.0875 & 1.0053 & 0.9168 \\
 & GraNd & 5.3343 & 2.9467 & 2.1475 & 1.8400 & 1.6018 & 1.3504 & 1.1566 & 1.0438 & 1.0004 & 0.9432 \\
 & MetaBench & 2.7413 & 2.0771 & 1.6838 & 1.4471 & 1.2477 & 1.1103 & 0.9767 & 0.9493 & 0.8626 & 0.7490 \\
 & \cellcolor{green!5}\mymethod{} & \cellcolor{green!5}\textbf{2.3990} & \cellcolor{green!5}\textbf{1.4293} & \cellcolor{green!5}\textbf{1.1653} & \cellcolor{green!5}\textbf{0.8023} & \cellcolor{green!5}\textbf{0.7192} & \cellcolor{green!5}\textbf{0.6045} & \cellcolor{green!5}\textbf{0.5053} & \cellcolor{green!5}\textbf{0.4803} & \cellcolor{green!5}\textbf{0.4326} & \cellcolor{green!5}\textbf{0.3699} \\ \midrule
\multirow{5}{*}{HellaSwag} & Random & 2.5738 & 1.8071 & 1.4658 & 1.1984 & 1.1569 & 1.0196 & 0.9751 & 0.8933 & 0.7973 & 0.8342 \\ 
 & PPL & 2.9751 & 1.9949 & 1.6428 & 1.2980 & 1.0798 & 0.9306 & 0.9162 & 0.8333 & 0.7944 & 0.7466 \\
 & GraNd & 2.9223 & 2.0102 & 1.5892 & 1.2572 & 1.0425 & 0.9603 & 0.8844 & 0.8241 & 0.8423 & 0.7825 \\
 & MetaBench & 2.4339 & 1.6940 & 1.4675 & 1.3135 & 1.1683 & 1.0668 & 0.9926 & 0.9120 & 0.8707 & 0.8220 \\
 & \cellcolor{green!5}\mymethod{} & \cellcolor{green!5}\textbf{2.2639} & \cellcolor{green!5}\textbf{1.5717} & \cellcolor{green!5}\textbf{1.2323} & \cellcolor{green!5}\textbf{1.0638} & \cellcolor{green!5}\textbf{0.8906} & \cellcolor{green!5}\textbf{0.7483} & \cellcolor{green!5}\textbf{0.6679} & \cellcolor{green!5}\textbf{0.6150} & \cellcolor{green!5}\textbf{0.5332} & \cellcolor{green!5}\textbf{0.5111} \\ \midrule
\multirow{5}{*}{WinoGrande} & Random & 3.4995 & 2.8713 & 2.1483 & 1.9490 & 1.5938 & 1.5314 & 1.2768 & 1.1537 & 1.1279 & 0.9854 \\
 & PPL & 4.2685 & 2.7479 & 2.3403 & 1.9352 & 1.7909 & 1.7748 & 1.5706 & 1.4714 & 1.3994 & 1.3176 \\
 & GraNd & 4.2685 & 2.6562 & 2.3045 & 2.0138 & 1.7775 & 1.7665 & 1.6155 & 1.5049 & 1.4037 & 1.2726 \\
 & MetaBench & 2.7834 & 2.1219 & 1.7515 & 1.5297 & 1.2893 & 1.2030 & 1.0722 & 0.9578 & 0.8658 & 0.7850 \\
 & \cellcolor{green!5}\mymethod{} & \cellcolor{green!5}\textbf{2.5086} & \cellcolor{green!5}\textbf{1.3994} & \cellcolor{green!5}\textbf{0.9791} & \cellcolor{green!5}\textbf{0.7772} & \cellcolor{green!5}\textbf{0.6307} & \cellcolor{green!5}\textbf{0.5580} & \cellcolor{green!5}\textbf{0.5098} & \cellcolor{green!5}\textbf{0.4521} & \cellcolor{green!5}\textbf{0.4134} & \cellcolor{green!5}\textbf{0.3905} \\ \midrule
\multirow{5}{*}{MMLU} & Random & 3.5048 & 2.2881 & 2.1036 & 1.9096 & 1.5779 & 1.5901 & 1.4984 & 1.3357 & 1.3357 & 1.1699 \\
 & PPL & 8.0290 & 9.7627 & 10.4998 & 8.5047 & 8.0146 & 7.8817 & 7.6781 & 7.2748 & 7.2060 & 6.7814 \\
 & GraNd & 8.9996 & 10.0913 & 10.4750 & 8.7332 & 8.1563 & 7.9034 & 7.7453 & 7.2225 & 7.5776 & 6.8507 \\
 & MetaBench & 2.4268 & 2.0925 & 1.7382 & 1.5292 & 1.3617 & 1.2872 & 1.1992 & 1.1401 & 1.0626 & 0.9941 \\
 & \cellcolor{green!5}\mymethod{} & \cellcolor{green!5}\textbf{2.4117} &  \cellcolor{green!5}\textbf{1.8293} & \cellcolor{green!5}\textbf{1.3951} & \cellcolor{green!5}\textbf{1.1126} & \cellcolor{green!5}\textbf{1.0220} & \cellcolor{green!5}\textbf{0.8460} & \cellcolor{green!5}\textbf{0.7667} & \cellcolor{green!5}\textbf{0.6906} & \cellcolor{green!5}\textbf{0.6406} & \cellcolor{green!5}\textbf{0.5966} \\ \bottomrule
\end{tabular}
}
\vspace{-20pt}
\end{table*}

\noindent \textbf{Mutation}. In this process, the aim is to introduce randomness into the new individual $\mathbf{m}^{(c)}$, let $\lambda_j$ be a flag which sample of an individual should mutate: 
\begin{equation}
\setlength{\abovedisplayskip}{3pt}
\setlength{\belowdisplayskip}{3pt}
m^{(c)}_j \gets m^{(c)}_j \oplus \lambda_j, \end{equation}
where $\lambda_j \sim \text{Bernoulli}\left(\frac{1}{k}\right),\; j \in [1, M]$.

\noindent \textbf{Adjustment}. To guarantee \( \mathbf{m}^{(c)} \) has $k$-ones, the \textit{adjustment} process is implemented by randomly setting superfluous ones to zeros, vice versa.


\noindent{\textbf{Step 3: Attribution-based Sample Selection}}.\label{subsubsec: div} To maximize reconstruction fidelity while preserving representational diversity during dataset compression, we train an Explainable Boosting Machine (EBM)~\citep{nori2019interpretml} on the elite mask set $\mathcal{E} = \{\mathbf{m}^{(1)},\dots,\mathbf{m}^{(N_{\mathcal{E}})}\}$ from Step 2, assigning each sample in $\mathcal{D}_{\mathrm{filtered}} = \{x_1,\dots,x_M\}$ a data-specific attribution score; these scores are used to stratify samples into groups, upon which a genetic algorithm (GA) performs optimized selection to balance signal strength and coverage. The result is a compressed dataset $\mathcal{D}_{\mathrm{compressed}}$ of size $P < M$, which retains high-impact instances while preserving underrepresented patterns — achieving efficient, robust, and generalizable compression without compromising reconstruction performance.

To quantify how much each sample contributes to reconstruction accuracy in the predictor model, we first define the \textit{attribution} of each sample within a mask. For a mask $\mathbf{m}\in\mathcal{E}$, define the selected index set
$\mathcal{I}(\mathbf{m})=\{j\,|\,m_{j}=1\}$.
An EBM
$
g_{\mathbf{m}}(\mathcal{D_{\mathrm{filtered}}})
      =\sum_{j\in\mathcal{F}(\mathbf{m})}f^{\mathbf{m}}_{j}(x_j)
$
learns the training data $\mathcal{T'}$. Let $\mathbf{S}_{\mathrm{filtered},i}$ denote the $i$-th row of $\mathbf{S}_{\mathrm{filtered}}$, then the form of $\mathcal{T'}$ is: \(\{\mathbf{S}_{\mathrm{filtered},i},y_i\}_{i\in\mathcal{T'}}\), which constructs a map from sample score matrix to whole dataset accuracy.
The component norm  
$\|f^{\mathbf{m}}_{j}\|_{2}$
in $g_\mathbf{m}$ is defined as the \textit{attribution} of sample~$j$ in a mask.
Aggregating over all attributions of each mask in $\mathcal{E}$ yields the \emph{global attribution} of sample~$j$, denoted as $A_j$:
\begin{equation}
\setlength{\abovedisplayskip}{3pt}
\setlength{\belowdisplayskip}{3pt}
A_j
=\;
\frac{%
  \sum_{\mathbf{m}\in\mathcal{E}}
    \mathbf 1\{\,j\in\mathcal{I}(\mathbf{m})\}\,\|f^{\mathbf{m}}_{j}\|_{2}%
}{%
  \sum_{\mathbf{m}\in\mathcal{E}}
    \mathbf 1\{\,j\in\mathcal{I}(\mathbf{m})\}%
},
\label{eq:imp_mean}
\end{equation}
where $j\in\mathcal{I}(\mathcal{D}_{\mathrm{filtered}})
$. According to the attributions, a tri-partition of the samples can be implemented. With a retention ratio $\alpha\in(0,1)$, set $q=\bigl\lceil\alpha M\bigr\rceil=P\ll M$ and create three groups of \emph{equal} size~$q$: $G_{\mathrm{high}},G_{\mathrm{low}},G_{\mathrm{rand}},$where $G_{\mathrm{high}}$ contains the $q$ samples with the
largest $A_j$, $G_{\mathrm{low}}$ the smallest ones, and $G_{\mathrm{rand}}$ the random ones. The grouping operation deliberately forces subsequent GAs to search in regions of different attributions so that information neglected by the current top-$N_{\mathcal{E}}$ subsets can be rediscovered, while the information that is really significant can be boosted. For every group $G\in \{G_{\mathrm{high}}, G_{\mathrm{low}},G_{\mathrm{rand}}\}$, GA is used to judge the best group to get $\mathcal{D_{\mathrm{compressed}}}$. 

To increase diversity of pruned dataset, we iteratively repeat Step 2 and Step 3, at each round the globally best mask
$\mathbf{m}^{\star}$
is updated whenever a lower error is observed. The details of the entire process are provided in Algorithm~\ref{alg:ga_step2} in Appendix ~\ref{sec:Pseudo Code}.



\section{Experiments}
\subsection{Experimental Setup}
\textbf{Baseline Methods}. 
We compared our method with other benchmark compression approaches, including MetaBench. In addition, we also conducted comparisons with several classic methods, including Random Selection, GraNd~\citep{paul2021deep}, and Perplexity (PPL)~\citep{ppl}. 
The total score of all selected subsets from a given dataset is predicted using a GAM trained on the dataset, and RMSE is computed. The subset with the lowest RMSE is selected as the optimal subset. Please refer to Appendix ~\ref{appendix:baseline_method} for more details.

\noindent\textbf{Dataset Construction}.
We constructed our dataset using data from the \texttt{Open LLM Leaderboard}~\citep{OpenLLMLeaderBoard} and conducted extensive evaluations of our method and the baselines. The datasets include \texttt{GSM8K}~\citep{GSM8K} (1K samples), \texttt{ARC}~\citep{ARC} (400 samples), \texttt{HellaSwag}~\citep{HelloSwag} (10K samples), \texttt{WinoGrande}~\citep{winogrande} (44K samples), and \texttt{MMLU}~\citep{MMLU} (15K samples). For data preprocessing, the protocols proposed in MetaBench \citep{MetaBench} were adopted, which involved the removal of low-performing models and items with low variance. For both the training and testing sets, the dataset is first ranked by score and then partitioned into ten equipotent strata. Within each stratum, 10\% of the instances are randomly sampled and subsequently pooled to constitute the test set; the remaining 90\% are retained as the training set. Please refer to the Appendix ~\ref{appendix:dataset_construction} for more details.
\subsection{Main Results}
\noindent\textbf{Better Performance with the Same Compression Ratios}. For the results on five benchmarks in Table~\ref{tab:rmse}, compared with previous methods, \mymethod{} achieves state-of-the-art (SOTA) results across all datasets. Notably, on \texttt{GSM8K} with a subset size of 500, \mymethod{} achieves a 60.7\% reduction in RMSE compared to MetaBench, demonstrating its consistently superior capability to preserve performance under highly constrained data regimes.

\noindent\textbf{Comparable Performance with Smaller Compression Ratios}.
\mymethod{} achieves comparable or superior performance while using significantly smaller subset sizes, indicating improved data efficiency. For instance, on \texttt{GSM8K}, \mymethod{} surpasses the performance of MetaBench using only 200 examples, whereas MetaBench requires 500. A similar trend is observed on the WinoGrande dataset, where EssenceBench outperforms MetaBench with the same reduced subset size of 200, again highlighting its effectiveness under tighter data budgets.

\subsection{Ablation Study}
\noindent\textbf{Parameter Sensitivity Analysis}.
To evaluate how hyperparameters influence the effectiveness of \mymethod{}, we analyze two factors: (1) the number of generations (\textit{gens}) used in the genetic algorithm to evolve candidate subsets within each iteration, and (2) the number of iterative refinement rounds combining subset search (Step 2) and attribution-based grouping (Step 3), as shown in Figure~\ref{fig:pipeline}. 
\begin{wraptable}[7]{r}{0.5\linewidth} 
    \centering
    \vspace{-10pt} 
    \caption{Performance over Generations and Rouns.}
    \label{tab:args}
    \vspace{-5pt}
    \resizebox{\linewidth}{!}{%
    \begin{tabular}{c|*{4}{>{\centering\arraybackslash}p{1.5cm}}}
        \toprule 
        \multirow{2}{*}{Gen} & \multicolumn{4}{c}{Round}\\
         & 2 & 3 & 4 & 5 \\
        \midrule
        1000  & 2.7685 & 2.8940 & 2.7050 & 2.4731 \\
        2000  & 2.8295 & 2.7727 & 2.5338 & 2.4186 \\
        3000  & 3.0065 & 2.7494 & 2.5900 & 2.4015 \\
        \bottomrule
    \end{tabular}
    }
    \vspace{-20pt} 
\end{wraptable}
This analysis aims to reveal the tradeoff between search depth and multi-round refinement in achieving low reconstruction error. Experimental results in Table~\ref{tab:args} show that increasing the number of refinement rounds consistently improves performance, regardless of \textit{gens}. For example, with \textit{gens} fixed at 1000, extending from 2 to 5 rounds reduces RMSE from 2.77 to 2.47. When rounds are sufficient (e.g. 5), higher \textit{gens} provides steady improvements, demonstrating that deeper search becomes valuable only when paired with adequate downstream selection. This empirical observation confirms the consistent effectiveness of repeated attribution-guided filtering and recompression in progressively reducing error. On the other hand, increasing \textit{gens} under a small number of rounds (e.g. 2) yields marginal gains or even degrades performance, likely due to over-exploration in a limited refinement context. All results are reported on \texttt{GSM8K} using a fixed 50-sample training set.

\begin{figure}[tb!]
    \centering
    \includegraphics[width=\linewidth]{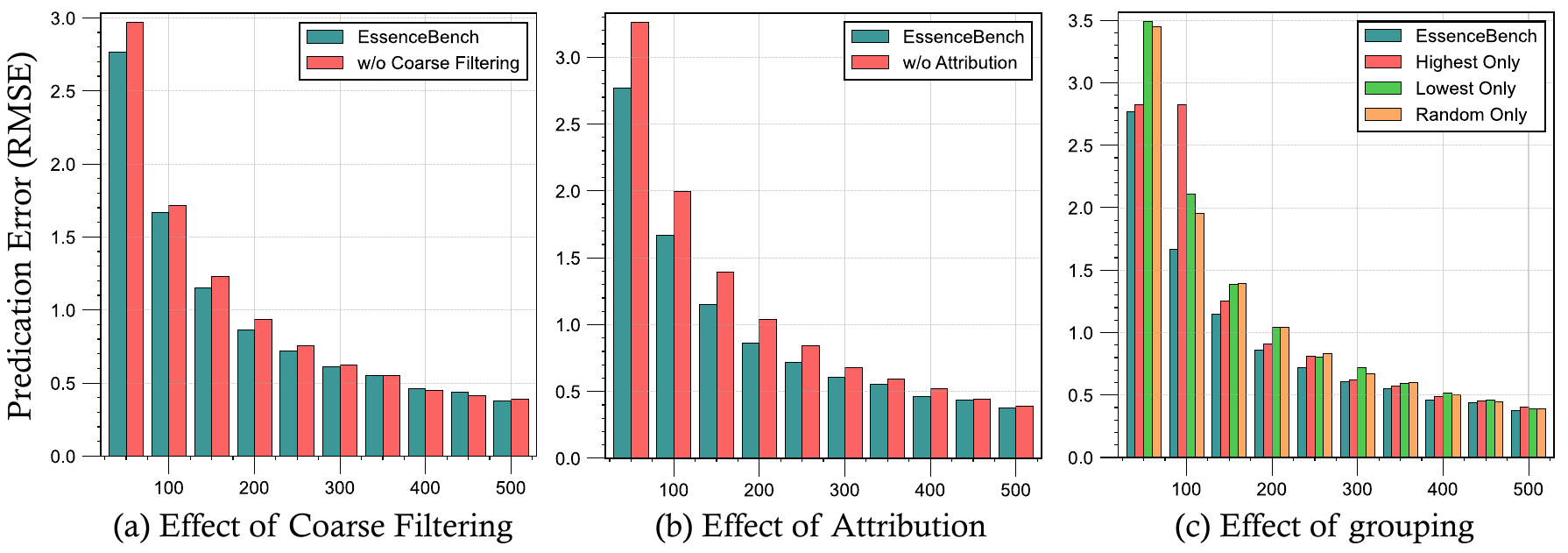}
    \vspace{-10pt}
    \caption{Ablation results on \texttt{GSM8K}, evaluating the effect of (a) coarse filtering, (b) attribution-based selection, and (c) grouping strategies.}
    \label{fig:ablation}
    \vspace{-20pt}
\end{figure}

\noindent\textbf{The impact of coarse filtering}.
To isolate the effect of coarse filtering, we compare the full \mymethod{} pipeline with a variant that skips redundancy-based filtering and retains only basic outlier removal (Raw \mymethod{}). As shown in Figure~\ref{fig:ablation}(a), applying coarse filtering significantly reduces RMSE, confirming its effectiveness in removing redundant samples that otherwise degrade reconstruction quality. This performance benefit is most pronounced especially when the subset size is small. As the subset grows, this gain diminishes, likely because the genetic algorithm is more likely to include informative examples regardless of filtering.

\noindent\textbf{The impact of attribution}.
To investigate the impact of attribution, which is the base of sample selection (Step 3), we compare \mymethod{} with merely GA without attribution-based sample selection. As shown in Figure~\ref{fig:ablation}(b), when the subset size is below 400, attribution-based sample selection outperforms selection without attribution. However, this notable performance advantage largely disappears once the subset size exceeds 400.

\noindent\textbf{The impact of grouping}.
To assess the role of grouping in Step 3, we compare \mymethod{} against three variants: selecting only the top-attribution group (Highest Only), the lowest-attribution group (Lowest Only), or a randomly sampled group (Random Only). All experiments are conducted on \texttt{GSM8K}. As shown in Figure~\ref{fig:ablation}(c), \mymethod{} consistently outperforms the alternatives when the subset size is small, indicating that attribution-guided diversity effectively enhances compression quality. However, when the subset size exceeds 400, all grouping strategies yield similar RMSE, suggesting diminishing returns from fine-grained selection under larger data budgets.


\subsection{Case Study}
\noindent \textbf{Text and Ranking Redundancy}. To assess the effectiveness of the coarse filtering strategy, we present representative samples with high text and ranking redundancy.
\begin{wraptable}{r}{0.55\textwidth} 
    \vspace{-5pt} 
    \caption{The most similar text (left) and the highest-ranked similar text (right).}
    \vspace{-5pt} 
    \label{fig:similar-text}
    \centering
    \renewcommand{\arraystretch}{2.0} 
    \large
    \resizebox{\linewidth}{!}{
    \begin{tabular}{p{5cm}|p{8cm}}  
        \hline
        \textcolor{YellowOrange}{Zack's locker is half as big as Timothy's locker. Peter's locker is 1/4 as big as Zack's locker.} \textcolor{ForestGreen}{If \textcolor{Red}{Peter's} locker is \textcolor{Red}{5} cubic inches,} \textcolor{blue}{how big is \textcolor{Red}{Timothy's} locker in cubic inches? }
        & 
        \textcolor{YellowOrange}{Axel has 50 silver pesos and 80 gold pesos.} \textcolor{blue}{He visits her friend Anna who has twice as many silver pesos as he has and 40 more gold pesos.} What's \textbf{the total number} of pesos they have together? 
        \\ \hline
        \textcolor{ForestGreen}{\textcolor{Red}{Timothy's} locker is \textcolor{Red}{24} cubic inches.} \textcolor{YellowOrange}{Zack's locker is half as big as Timothy's locker.Peter's locker is 1/4 as big as Zack's locker.} \textcolor{blue}{How big is \textcolor{Red}{Peter's} locker in cubic inches? }
        & 
        Amy is taking a history test. \textcolor{YellowOrange}{She correctly answers 80\% of the multiple-choice questions, 90\% of the true/false questions, and 60\% of the long-answer questions.} \textcolor{blue}{The multiple-choice and true/false questions are worth 1 point each, and the long answer questions are worth 5 points each.} How many \textbf{points does Amy score} if there are 10 multiple-choice questions, 20 true/false questions, and 5 long answer questions? 
        \\ \hline
    \end{tabular}
    }
    \vspace{-10pt} 
\end{wraptable}
 As clearly illustrated in Table~\ref{fig:similar-text}, the proposed filtering mechanism successfully and consistently identifies semantically equivalent items across various cases. In the text redundancy case, two questions differ in phrasing but share the same arithmetic structure. In the ranking redundancy case, two problems with different narratives yield similar model scores because both problems require multi-step numerical reasoning that includes proportional calculations, intermediate variable derivation, and aggregation of weighted quantities. These representative cases provide clear evidence that \mymethod{} effectively captures and distinguishes both superficial and deeper structural redundancies.

\noindent \textbf{Ranking Distribution}. To comprehensively evaluate how well the predicted accuracies preserve the original model rankings, we computed several ranking metrics, detailed in the Appendix~\ref{sec:rank_distribution} and Tables~\ref{tab:gsm8k}, \ref{tab:hellaswag}, and \ref{tab:arc}. In particular, we primarily focus on the \textit{ranking error} metric (also referred to as \textit{ranking changing}), which measures the proportion of models whose predicted rank deviates from their true rank by no more than a specified percentage of the total number of models (e.g., within 5\% or 10\% rank positions). Notably, we found that selecting just 200$\times$ less samples preserves 95\% of rankings changing within 10\%. Furthermore, as shown in Figure~\ref{fig:ranking_case}, \mymethod{} consistently outperforms MetaBench in maintaining ranking fidelity. With only 200 samples, all ranking shifts remain within 10\%, and with 400 samples, within 5\%, offering significantly tighter preservation compared to the deviations observed under MetaBench. 
\begin{figure}[tb!]
    \centering
    \vspace{-15pt} \includegraphics[width=.90\linewidth]{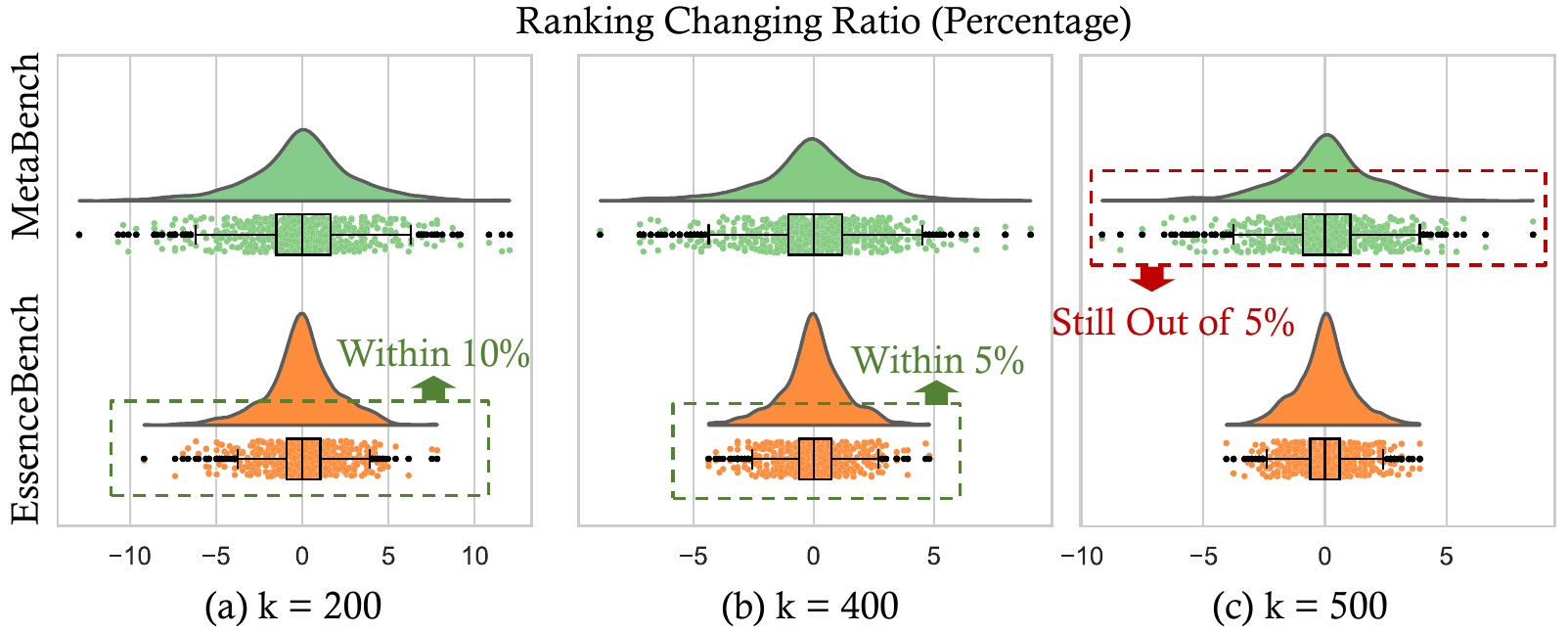}
    \vspace{-10pt}
    \caption{Comparison of ranking change distributions between MetaBench and \mymethod{} on the \texttt{HellaSwag} dataset, where $k$ denotes the subset size.}
    \vspace{-20pt}
    \label{fig:ranking_case}
\end{figure}


\section{Conclusion}
In this paper, we identified sample redundancy in LLM benchmark evaluation. To address this problem, we introduced \mymethod{}, a coarse-to-fine benchmark compression framework that combines redundancy-aware filtering with an iterative genetic algorithm optimized for accurate reconstruction. Extensive experiments on five standard benchmarks 
show that \mymethod{} achieves a 200$\times$ reduction in benchmark size while preserving ranking fidelity. 



\clearpage
\bibliography{iclr2026_conference}
\bibliographystyle{iclr2026_conference}
\clearpage
\appendix
\appendix
\section{Additional Related Works} \label{sec:more_related works}

To contextualize our benchmark compression setting within the broader literature, we briefly supplement here the main lines of work on data selection for large language models (LLMs).

Existing methods can be roughly divided into two categories. The first focuses on large-scale data filtering, aiming to eliminate low-quality, toxic, or duplicated samples from pretraining corpora~\citep{raffel2020exploring, marion2023less, zhang2022opt, abbas2023semdedup}. These approaches typically prioritize data cleanliness and safety rather than sample representativeness.

The second category explores subset selection strategies that retain model performance with fewer examples. Notably, several works leverage heuristics or Item Response Theory (IRT)-based scoring to select informative subsets~\citep{zhou2023lima, wang2023far, ivison2023camels}. More recent studies also explore the use of LLMs themselves to score or rank data~\citep{xia2024less, liu2023makes}, though most of these focus on task-specific learning efficiency rather than benchmark-level score reconstruction or ranking consistency.

Our setting differs in that we explicitly aim to preserve evaluation integrity across LLMs under compression, a goal not typically prioritized in prior work. This highlights the unique focus of our proposed method in balancing compression with benchmarking fidelity.

\section{Pseudo Code of \mymethod{}} \label{sec:Pseudo Code}
To enhance the reproducibility and clarity of our method, we provide detailed pseudo-code for the core components of \mymethod{}. The algorithm consists of two main stages: a single-round Genetic Algorithm for subset selection (Algorithm~\ref{alg:ga_step2}), and a multi-round coarse-to-fine compression framework that iteratively refines sample selection through attribution-based grouping (Algorithm~\ref{alg:ga_ebm}).

\paragraph{Algorithm~\ref{alg:ga_step2}: Genetic Algorithm for Subset Selection.}
This module performs evolutionary search over binary masks that represent subsets of size $k$ from the filtered benchmark. In each generation, the population undergoes fitness evaluation based on RMSE error, followed by tournament selection, crossover, mutation, and adjustment. The top-performing subsets (elites) are retained for both optimization and attribution calculation.
\begin{algorithm}[tb!]
\caption{Genetic Algorithm (Subset Selection)}
\label{alg:ga_step2}
\begin{algorithmic}[1]
\REQUIRE $\mathbf{S}_{\mathrm{filtered}}\in\{0,1\}^{N_{\mathrm{LLM}}\times M},\;\mathbf{y}\in\mathbb{R}^{n_{\mathrm{LLM}}},$ $k,\;N_{\mathcal{P}},\;N_\mathcal{E},\;N_G,\;g(\cdot)$
\ENSURE Best mask $\mathbf{m}^*$, error$\varepsilon^*$, and final top E subsets $\mathcal{E}$
\STATE Initialize population $\mathcal{P}=\{\mathbf{m}^{(i)}\}_{i=1}^{N_{\mathcal{P}}}$ with random $k$-masks
\STATE $\mathbf{m}^*\gets \mathbf{0},\;\varepsilon^*\gets +\infty$
\FOR{$t=1$ to $N_G$}
  \FOR{each $\mathbf{m}\in\mathcal{P}$}
    \STATE Compute accuracy $\hat s_i = \frac{1}{k}\mathbf{S}_{\mathrm{filtered}}\mathbf{m}$
    \STATE Predict $\hat y = g(\hat s_i)$
    \STATE Compute error $\varepsilon(\mathbf{m})
      = \sqrt{\frac{1}{|\mathcal{V}|}\sum_{i\in\mathcal{V}}(\hat y_i - y_i)^2}$
    \STATE Set fitness $F(\mathbf{m}) = -\,\varepsilon(\mathbf{m})$
  \ENDFOR
  \STATE Select the top $N_{\mathcal{E}}$ masks by fitness into $\mathcal{E}$
  \STATE $\mathcal{P}' \gets \mathcal{E}$
  \WHILE{$|\mathcal{P}'| < N_{\mathcal{P}}$}
    \STATE $\mathbf{m}^{(a)},\mathbf{p}^{(b)} \gets \text{tournament}(\mathcal{P})$
    \STATE $\mathbf{m}^{(c)} \gets \text{crossover}(\mathbf{m}^{(a)},\mathbf{m}^{(b)})$
    \STATE $\mathbf{m}^{(c)} \gets \text{mutate}(\mathbf{m}^{(c)})$
    \STATE Adjust $\mathbf{m}^{(c)}$ to have exactly $k$ ones
    \STATE $\mathcal{P}' \gets \mathcal{P}' \,\cup\,\{\mathbf{m}^{(c)}\}$
  \ENDWHILE
  \STATE $\mathcal{P} \gets \mathcal{P}'$
  \IF{$\min_{\mathbf{m}\in\mathcal{P}}\varepsilon(\mathbf{m}) < \varepsilon^*$}
    \STATE $\varepsilon^* \gets \min_{\mathbf{m}}\varepsilon(\mathbf{m})$
    \STATE $\mathbf{m}^* \gets \arg\min_{\mathbf{m}}\varepsilon(\mathbf{m})$
  \ENDIF
\ENDFOR
\STATE \RETURN $\mathbf{m}^*,\,\varepsilon^*,\,\mathcal{E}$
\end{algorithmic}
\end{algorithm}

\paragraph{Algorithm~\ref{alg:ga_ebm}: Iterative GA with Attribution-Guided Refinement.}
Building on Algorithm~\ref{alg:ga_step2}, this procedure incorporates sample-level attributions to enhance selection diversity and convergence. In each round, attribution scores are computed from the top-$N_{\mathcal{E}}$ elites using an Explainable Boosting Machine (EBM). Samples are then partitioned into three equally sized groups—\texttt{High}, \texttt{Low}, and \texttt{Random}—and GA is applied to each. The group achieving the lowest error becomes the new candidate pool for the next iteration. The process repeats until convergence or until the size of the pool falls below the desired subset size $k$.

\paragraph{Key Notations.}
We briefly summarize the main symbols used throughout our method. The score matrix after coarse filtering is denoted by $\mathbf{S}{\mathrm{filtered}}$, while $\mathbf{y}$ represents the ground-truth accuracies of all LLMs on the full benchmark. The target coreset size is given by $k$. In the genetic algorithm (GA) procedure, $N{\mathcal{P}}$ refers to the population size, and $N_{\mathcal{E}}$ denotes the number of top-performing (elite) candidates retained in each generation. Each GA round runs for $N_G$ generations. The outer coarse-to-fine loop terminates after at most $R_{\max}$ iterations. During attribution-based grouping, $\alpha$ controls the proportion of samples retained in each candidate group, and $\beta$ denotes the sampling temperature that introduces stochasticity in Step 3.

\begin{algorithm}[tb!]
\caption{\textsc{Iterative GA ( Subset Selection + Sample Selection)}}
\label{alg:ga_ebm}
\begin{algorithmic}[1]
\REQUIRE
  $\mathbf{S_{\mathrm{filtered}}}\!\in\!\{0,1\}^{n_{\mathrm{LLM}}\times M}$,\,$\mathbf{y}\in\mathbb{R}^{n_{\mathrm{LLM}}\times1}$,   $k$,
  $R_{\max}$,  $\alpha$,  $\beta$
\ENSURE
  best mask $\mathbf{m}^{\star}$,  error $\varepsilon^{\star}$
\STATE $\mathcal{I}\!\gets\!\{1,\dots,M\},\;
       \mathbf{m}^{\star}\!\gets\mathbf{0},\;
       \varepsilon^{\star}\!\gets\infty$
\FOR{$r=0$ \textbf{to} $R_{\max}-1$}
    \STATE $(\mathbf{m},\varepsilon,\mathcal{E})
           \gets\textsc{GA\_Search}\bigl(\mathbf{S}(:,\mathcal{I}),\mathbf{y},k\bigr)$
    \IF{$\varepsilon<\varepsilon^{\star}$}
        \STATE update $(\mathbf{m}^{\star},\varepsilon^{\star})$
    \ENDIF
    \IF{$|\mathcal{I}|\!\le\!k$ \textbf{or} $r=R_{\max}-1$}
        \STATE \textbf{break}
    \ENDIF
    \COMMENT{\emph{attributions aggregation}}
    \STATE compute attributions $\{A_j\}_{j\in\mathcal{I}}$ from the top-$N_\mathcal{E}$ subsets $\mathcal{E}$
    \STATE sort $A_j$; let $q=\lceil\alpha\cdot|\mathcal{I}|\rceil$
    \STATE $G_{\mathrm{high}}\!\gets$ top-$q$ samples,\;
           $G_{\mathrm{low}}\!\gets$ bottom-$q$ samples,\;
           $G_{\mathrm{rand}}\!\gets$ random-$q$ samples
    \FOR{$G\in\{G_{\text{high}},G_{\text{low}},G_{\text{rand}}\}$}
        \STATE $(\_,\varepsilon_{G},\_)\gets
               \textsc{GA\_Search}\bigl(\mathbf{S}(:,G),\mathbf{y},
               k\,;\ \beta\bigr)$
    \ENDFOR
    \STATE $\mathcal{I}\!\gets\!\arg\min_{G}\varepsilon_{G}$
\ENDFOR
\STATE \RETURN $\mathbf{m}^{\star},\varepsilon^{\star}$
\end{algorithmic}
\end{algorithm}

\clearpage

\section{Details of Ranking Distribution}
\label{sec:rank_distribution}
\paragraph{Metrics.}
Each table compares MetaBench and \mymethod{} on eight well-defined statistics that together quantify how faithfully a small subset reproduces the full leaderboard.

\noindent\textbf{RMSE ($\downarrow$).}
Let $y_i$ be the true accuracy of model $i$ on the full benchmark, and $\hat y_i$ the accuracy estimated from the compressed subset. With $n$ models,
\[
\mathrm{RMSE}= \sqrt{\frac{1}{n}\sum_{i=1}^{n}(y_i-\hat y_i)^2},
\]
so a lower value indicates more accurate prediction of overall scores.

\noindent\textbf{Rank correlations ($\uparrow$).}
We report three correlation metrics: Pearson’s $r$, Spearman’s $\rho$, and Kendall’s $\tau$. Pearson correlation is computed on raw scores $(y_i)$ and $(\hat y_i)$, reflecting linear agreement. Spearman and Kendall are based on rankings $\mathrm{rank}(y_i)$ and $\mathrm{rank}(\hat y_i)$, capturing monotonic consistency even when score magnitudes differ.

\noindent\textbf{Rank Stability ($\uparrow$).}
We define average positional deviation $\delta_i = |\mathrm{rank}(y_i) - \mathrm{rank}(\hat y_i)|$, normalised over all models:
\[
\mathrm{Stability} = 1 - \frac{1}{n}\sum_{i=1}^{n} \frac{\delta_i}{n}.
\]
This ranges from 1 (perfect ranking match) to 0 (completely disordered).

\noindent\textbf{Pair Accuracy ($\uparrow$).}
This measures the fraction of model pairs that preserve their relative ranking:
\[
\mathrm{PairAcc} = \frac{1}{\binom{n}{2}} \sum_{i<j} \mathbf{1}[(y_i > y_j) \Leftrightarrow (\hat y_i > \hat y_j)].
\]

\noindent\textbf{Top-tier retrieval ($\uparrow$).}
NDCG@50 is the Normalized Discounted Cumulative Gain over the top-50 predicted models, computed using the ground-truth order as ideal ranking. Top-50 Accuracy measures the intersection-over-union between true and predicted top-50 sets.

\noindent\textbf{Ranking Error within \{1,2,5,10\}\% ($\uparrow$).}
For a given tolerance $p \in \{1, 2, 5, 10\}$, this reports the proportion of models whose predicted rank deviates from the ground-truth rank by at most $\lceil pn/100 \rceil$ positions. Higher values mean better rank preservation under tighter constraints.

Together, these metrics assess both absolute score accuracy and ranking quality, from overall correlation down to local ordering and top-model retrieval fidelity.


\paragraph{Table organisation.}
Each block of rows corresponds to the metrics just defined, while the columns enumerate subset sizes from 50 to 500 test items. Within every cell the lower (for RMSE) or higher (for the seven ranking metrics) value is emboldened so that the superior method-MetaBench or \mymethod{}-is visible at a glance.

\paragraph{Key findings.}
On all three representative benchmarks \mymethod{} achieves the lowest RMSE, with the margin especially large when only 50–250 examples are retained. Correlation metrics (Pearson, Spearman, Kendall) and top-50 retrieval scores are likewise higher for \mymethod{}, indicating much closer alignment to the full-set leaderboard. In GSM8K and ARC, for example, \mymethod{} attains the same ranking stability with roughly 150–200 samples that MetaBench needs 400–500 samples to reach, underscoring the efficiency of our coarse-to-fine search.

Comprehensive numbers are reported in Tables \ref{tab:gsm8k}, \ref{tab:hellaswag} and \ref{tab:arc}.

\begin{table*}[tb!]
\centering
\renewcommand{\arraystretch}{1.2}
\small
\setlength{\tabcolsep}{6pt} 
\caption{MetaBench vs. EssenceBench on GSM8K dataset \footnotesize(↑ larger is better, ↓ smaller is better)}
\label{tab:gsm8k}
\resizebox{\textwidth}{!}{
\begin{tabular}{l|*{10}{r}}
\toprule
\multirow{2}{*}{\textbf{Method \& Metric}} & \multicolumn{10}{c}{\textbf{GSM8K Coreset Size}}\\
\cmidrule(l){2-11}&
\multicolumn{1}{c}{\;\;\;50}  & \multicolumn{1}{c}{\;\;\;100} & \multicolumn{1}{c}{\;\;\;150} &  \multicolumn{1}{c}{\;\;\;200} & \multicolumn{1}{c}{\;\;\;250} & \multicolumn{1}{c}{\;\;\;300} & \multicolumn{1}{c}{\;\;\;350} & \multicolumn{1}{c}{\;\;\;400} & \multicolumn{1}{c}{\;\;\;450} & \multicolumn{1}{c}{\;\;\;500} \\
\midrule

\rowcolor{metaBG}
\multicolumn{11}{l}{\textbf{MetaBench}}\\
RMSE↓               & 
\valmeta{3.508}     &
\valmeta{2.563}     &
\valmeta{2.053}     &
\valmeta{1.765}     &
\valmeta{1.535}     &
\valmeta{1.388}     &
\valmeta{1.209}     &
\valmeta{1.146}     &
\valmeta{1.047}     &
\valmeta{0.960}\\
Pearson↑            & 
\valmeta{0.991}&
\valmeta{0.995}&
\valmeta{0.997}&
\valmeta{0.998}&
\valmeta{0.998}&
\valmeta{0.999}&
\valmeta{0.999}&
\valmeta{0.999}&
\valmeta{0.999}&
\valmeta{0.999}\\
Spearman↑           & 
\valmeta{0.984}&
\valmeta{0.989}&
\valmeta{0.994}&
\valmeta{0.994}&
\valmeta{0.996}&
\valmeta{0.996}&
\valmeta{0.997}&
\valmeta{0.997}&
\valmeta{0.998}&
\valmeta{0.998}\\
Kendall↑            & 
\valmeta{0.888}&
\valmeta{0.912}&
\valmeta{0.934}&
\valmeta{0.937}&
\valmeta{0.949}&
\valmeta{0.950}&
\valmeta{0.956}&
\valmeta{0.960}&
\valmeta{0.964}&
\valmeta{0.967}\\
Rank Stability↑         & 
\valmeta{0.018}&
\valmeta{0.028}&
\valmeta{0.033}&
\valmeta{0.033}&
\valmeta{0.041}&
\valmeta{0.039}&
\valmeta{0.043}&
\valmeta{0.049}&
\valmeta{0.056}&
\valmeta{0.059}\\
Pair Accuracy↑          & 
\valmeta{0.930}&
\valmeta{0.948}&
\valmeta{0.960}&
\valmeta{0.963}&
\valmeta{0.970}&
\valmeta{0.971}&
\valmeta{0.975}&
\valmeta{0.977}&
\valmeta{0.979}&
\valmeta{0.981}\\
NDCG@50↑            & 
\valmeta{0.976}&
\valmeta{0.990}&
\valmeta{0.995}&
\valmeta{0.991}&
\valmeta{0.994}&
\valmeta{0.995}&
\valmeta{0.996}&
\valmeta{0.995}&
\valmeta{0.998}&
\valmeta{0.998}\\
Top50 Accuracy↑        & 
\valmeta{0.720}&
\valmeta{0.820}&
\valmeta{\textbf{0.880}}&
\valmeta{0.840}&
\valmeta{\textbf{0.880}}&
\valmeta{0.900}&
\valmeta{0.860}&
\valmeta{0.840}&
\valmeta{0.920}&
\valmeta{\textbf{0.940}}\\
Ranking Error within 1\%↑         & 
\valmeta{0.185}&
\valmeta{0.245}&\valmeta{0.298}&\valmeta{0.358}&\valmeta{0.386}&\valmeta{0.406}&\valmeta{0.488}&\valmeta{0.517}&\valmeta{0.504}&\valmeta{0.597}\\
Ranking Error within 2\%↑         & \valmeta{0.357}&\valmeta{0.457}&\valmeta{0.556}&\valmeta{0.596}&\valmeta{0.660}&\valmeta{0.681}&\valmeta{0.741}&\valmeta{0.755}&\valmeta{0.786}&\valmeta{0.841}\\
Ranking Error within 5\%↑         & \valmeta{0.684}&\valmeta{0.794}&\valmeta{0.887}&\valmeta{0.897}&\valmeta{0.941}&\valmeta{0.943}&\valmeta{0.953}&\valmeta{0.966}&\valmeta{0.984}&\valmeta{0.984}\\
Ranking Error within 10\%↑        & \valmeta{0.953}&\valmeta{0.959}&\valmeta{0.987}&\valmeta{0.984}&\valmeta{0.997}&\valmeta{0.993}&\valmeta{0.998}&\valmeta{0.998}&\valmeta{\textbf{1.000}}&\valmeta{\textbf{1.000}}\\

\midrule
\rowcolor{essBG}
\multicolumn{11}{l}{\textbf{EssenceBench}}\\
RMSE↓               & \valmeta{\textbf{2.900}} & \valmeta{\textbf{1.525}} & \valmeta{\textbf{1.131}} & \valmeta{\textbf{0.857}} & \valmeta{\textbf{0.682}} & \valmeta{\textbf{0.597}} & \valmeta{\textbf{0.543}} & \valmeta{\textbf{0.454}} & \valmeta{\textbf{0.426}} & \valmeta{\textbf{0.375}}\\
Pearson↑            & \valmeta{\textbf{0.994}} & \valmeta{\textbf{0.998}} & \valmeta{\textbf{0.999}} & \valmeta{\textbf{0.999}} & \valmeta{\textbf{1.000}} & \valmeta{\textbf{1.000}} & \valmeta{\textbf{1.000}} & \valmeta{\textbf{1.000}} & \valmeta{\textbf{1.000}} & \valmeta{\textbf{1.000}}\\
Spearman↑           & \valmeta{\textbf{0.987}} & \valmeta{\textbf{0.995}} & \valmeta{\textbf{0.997}} & \valmeta{\textbf{0.998}} & \valmeta{\textbf{0.999}} & \valmeta{\textbf{0.999}} & \valmeta{\textbf{0.999}} & \valmeta{\textbf{0.999}} & \valmeta{\textbf{0.999}} & \valmeta{\textbf{1.000}}\\
Kendall↑            & \valmeta{\textbf{0.905}} & \valmeta{\textbf{0.943}} & \valmeta{\textbf{0.955}} & \valmeta{\textbf{0.968}} & \valmeta{\textbf{0.973}} & \valmeta{\textbf{0.976}} & \valmeta{\textbf{0.978}} & \valmeta{\textbf{0.980}} & \valmeta{\textbf{0.982}} & \valmeta{\textbf{0.985}}\\
Rank Stability↑         & \valmeta{\textbf{0.023}}          & \valmeta{\textbf{0.033}}          & \valmeta{\textbf{0.046}}          & \valmeta{\textbf{0.074}}          & \valmeta{\textbf{0.097}}          & \valmeta{\textbf{0.093}}          & \valmeta{\textbf{0.100}}          & \valmeta{\textbf{0.116}}          & \valmeta{\textbf{0.128}}          & \valmeta{\textbf{0.165}}\\
Pair Accuracy↑          & \valmeta{\textbf{0.939}} & \valmeta{\textbf{0.963}} & \valmeta{\textbf{0.972}} & \valmeta{\textbf{0.979}} & \valmeta{\textbf{0.983}} & \valmeta{\textbf{0.984}} & \valmeta{\textbf{0.986}} & \valmeta{\textbf{0.987}} & \valmeta{\textbf{0.988}} & \valmeta{\textbf{0.990}}\\
NDCG@50↑            & \valmeta{\textbf{0.983}} & \valmeta{\textbf{0.993}} & \valmeta{\textbf{0.996}} & \valmeta{\textbf{0.998}} & \valmeta{\textbf{0.998}} & \valmeta{\textbf{0.999}} & \valmeta{\textbf{0.999}} & \valmeta{\textbf{1.000}} & \valmeta{\textbf{0.999}} & \valmeta{\textbf{1.000}}\\
Top50 Accuracy↑        & \valmeta{\textbf{0.780}} & \valmeta{\textbf{0.860}} & \valmeta{0.860}          & \valmeta{\textbf{0.940}} & \valmeta{\textbf{0.880}}          & \valmeta{\textbf{0.940}} & \valmeta{\textbf{0.900}} & \valmeta{\textbf{0.920}} & \valmeta{\textbf{0.940}} & \valmeta{0.920}\\
Ranking Error within 1\%↑         & \valmeta{\textbf{0.237}} & \valmeta{\textbf{0.383}} & \valmeta{\textbf{0.457}} & \valmeta{\textbf{0.609}} & \valmeta{\textbf{0.661}} & \valmeta{\textbf{0.727}} & \valmeta{\textbf{0.730}} & \valmeta{\textbf{0.792}} & \valmeta{\textbf{0.813}} & \valmeta{\textbf{0.864}}\\
Ranking Error within 2\%↑         & \valmeta{\textbf{0.416}} & \valmeta{\textbf{0.637}} & \valmeta{\textbf{0.727}} & \valmeta{\textbf{0.822}} & \valmeta{\textbf{0.879}} & \valmeta{\textbf{0.902}} & \valmeta{\textbf{0.912}} & \valmeta{\textbf{0.923}} & \valmeta{\textbf{0.944}} & \valmeta{\textbf{0.964}}\\
Ranking Error within 5\%↑         & \valmeta{\textbf{0.774}} & \valmeta{\textbf{0.917}} & \valmeta{\textbf{0.946}} & \valmeta{\textbf{0.982}} & \valmeta{\textbf{0.990}} & \valmeta{\textbf{0.990}} & \valmeta{\textbf{0.995}} & \valmeta{\textbf{0.997}} & \valmeta{\textbf{0.995}} & \valmeta{\textbf{1.000}}\\
Ranking Error within 10\%↑        & \valmeta{\textbf{0.956}} & \valmeta{\textbf{0.989}} & \valmeta{\textbf{1.000}} & \valmeta{\textbf{1.000}} & \valmeta{\textbf{1.000}} & \valmeta{\textbf{1.000}} & \valmeta{\textbf{1.000}} & \valmeta{\textbf{1.000}} & \valmeta{\textbf{1.000}} & \valmeta{\textbf{1.000}}\\
\bottomrule
\end{tabular}
}
\end{table*}

\begin{table*}[tb!]
\centering
\renewcommand{\arraystretch}{1.2}
\small
\setlength{\tabcolsep}{6pt} 
\caption{MetaBench vs. EssenceBench on HellaSwag dataset \footnotesize(↑ larger is better, ↓ smaller is better)}
\label{tab:hellaswag}
\resizebox{\textwidth}{!}{
\begin{tabular}{l|*{10}{r}}
\toprule
\multirow{2}{*}{\textbf{Method \& Metric}} & \multicolumn{10}{c}{\textbf{HellaSwag Coreset Size}}\\
\cmidrule(l){2-11}&
\multicolumn{1}{c}{\;\;\;50}  & \multicolumn{1}{c}{\;\;\;100} & \multicolumn{1}{c}{\;\;\;150} &  \multicolumn{1}{c}{\;\;\;200} & \multicolumn{1}{c}{\;\;\;250} & \multicolumn{1}{c}{\;\;\;300} & \multicolumn{1}{c}{\;\;\;350} & \multicolumn{1}{c}{\;\;\;400} & \multicolumn{1}{c}{\;\;\;450} & \multicolumn{1}{c}{\;\;\;500} \\

\rowcolor{metaBG}
\multicolumn{11}{l}{\textbf{MetaBench}}\\
RMSE↓            & \valmeta{2.397} & \valmeta{1.734} & \valmeta{1.495} & \valmeta{1.318} & \valmeta{1.166} & \valmeta{1.076} & \valmeta{0.961} & \valmeta{0.929} & \valmeta{0.880} & \valmeta{0.836} \\
Pearson↑         & \valmeta{0.992} & \valmeta{0.996} & \valmeta{0.997} & \valmeta{0.998} & \valmeta{0.998} & \valmeta{0.998} & \valmeta{0.999} & \valmeta{0.999} & \valmeta{0.999} & \valmeta{0.999} \\
Spearman↑        & \valmeta{0.979} & \valmeta{0.989} & \valmeta{0.992} & \valmeta{0.994} & \valmeta{0.995} & \valmeta{0.996} & \valmeta{0.997} & \valmeta{0.997} & \valmeta{0.997} & \valmeta{0.997} \\
Kendall↑         & \valmeta{0.879} & \valmeta{0.915} & \valmeta{0.928} & \valmeta{0.937} & \valmeta{0.942} & \valmeta{0.946} & \valmeta{0.955} & \valmeta{0.954} & \valmeta{0.957} & \valmeta{0.959} \\
Rank\_Stability↑      & \valmeta{0.024} & \valmeta{0.024} & \valmeta{0.021} & \valmeta{0.030} & \valmeta{0.045} & \valmeta{0.032} & \valmeta{0.054} & \valmeta{0.044} & \valmeta{0.062} & \valmeta{0.042} \\
Pair\_Accuracy↑       & \valmeta{0.921} & \valmeta{0.949} & \valmeta{0.957} & \valmeta{0.963} & \valmeta{0.967} & \valmeta{0.970} & \valmeta{0.974} & \valmeta{0.974} & \valmeta{0.976} & \valmeta{0.977} \\
NDCG@50↑         & \valmeta{0.990} & \valmeta{0.996} & \valmeta{\textbf{0.997}} & \valmeta{0.997} & \valmeta{0.997} & \valmeta{0.997} & \valmeta{0.998} & \valmeta{\textbf{0.999}} & \valmeta{\textbf{0.999}} & \valmeta{\textbf{0.999}} \\
Top50\_Accuracy↑     & \valmeta{0.920} & \valmeta{\textbf{0.940}} & \valmeta{\textbf{0.960}} & \valmeta{\textbf{0.940}} & \valmeta{0.900} & \valmeta{0.960} & \valmeta{0.900} & \valmeta{\textbf{0.960}} & \valmeta{0.940} & \valmeta{0.980} \\
Ranking\_Error\_within 1\%↑     & \valmeta{0.200} & \valmeta{0.265} & \valmeta{0.278} & \valmeta{0.355} & \valmeta{0.377} & \valmeta{0.395} & \valmeta{0.418} & \valmeta{0.435} & \valmeta{0.492} & \valmeta{0.466} \\
Ranking\_Error\_within 2\%↑       & \valmeta{0.370} & \valmeta{0.487} & \valmeta{0.537} & \valmeta{0.586} & \valmeta{0.621} & \valmeta{0.629} & \valmeta{0.705} & \valmeta{0.683} & \valmeta{0.717} & \valmeta{0.725} \\
Ranking\_Error\_within 5\%↑      & \valmeta{0.660} & \valmeta{0.812} & \valmeta{0.854} & \valmeta{0.889} & \valmeta{0.905} & \valmeta{0.928} & \valmeta{0.959} & \valmeta{0.949} & \valmeta{0.961} & \valmeta{0.974} \\
Ranking\_Error\_within 10\%↑       & \valmeta{0.902} & \valmeta{0.961} & \valmeta{0.983} & \valmeta{0.989} & \valmeta{0.991} & \valmeta{\textbf{1.000}} & \valmeta{0.998} & \valmeta{\textbf{1.000}} & \valmeta{0.998} & \valmeta{\textbf{1.000}} \\

\midrule
\rowcolor{essBG}
\multicolumn{11}{l}{\textbf{EssenceBench}}\\
RMSE↓            & \valmeta{\textbf{2.153}} & \valmeta{\textbf{1.453}} & \valmeta{\textbf{1.038}} & \valmeta{\textbf{0.863}} & \valmeta{\textbf{0.706}} & \valmeta{\textbf{0.635}} & \valmeta{\textbf{0.504}} & \valmeta{\textbf{0.461}} & \valmeta{\textbf{0.409}} & \valmeta{\textbf{0.419}} \\
Pearson↑         & \valmeta{\textbf{0.994}} & \valmeta{\textbf{0.997}} & \valmeta{\textbf{0.999}} & \valmeta{\textbf{0.999}} & \valmeta{\textbf{0.999}} & \valmeta{\textbf{0.999}} & \valmeta{\textbf{1.000}} & \valmeta{\textbf{1.000}} & \valmeta{\textbf{1.000}} & \valmeta{\textbf{1.000}} \\
Spearman↑        & \valmeta{\textbf{0.985}} & \valmeta{\textbf{0.993}} & \valmeta{\textbf{0.997}} & \valmeta{\textbf{0.997}} & \valmeta{\textbf{0.998}} & \valmeta{\textbf{0.998}} & \valmeta{\textbf{0.999}} & \valmeta{\textbf{0.999}} & \valmeta{\textbf{0.999}} & \valmeta{\textbf{0.999}} \\
Kendall↑         & \valmeta{\textbf{0.899}} & \valmeta{\textbf{0.931}} & \valmeta{\textbf{0.951}} & \valmeta{\textbf{0.959}} & \valmeta{\textbf{0.966}} & \valmeta{\textbf{0.967}} & \valmeta{\textbf{0.973}} & \valmeta{\textbf{0.974}} & \valmeta{\textbf{0.978}} & \valmeta{\textbf{0.978}} \\
Rank\_Stability↑      & \valmeta{\textbf{0.032}} & \valmeta{\textbf{0.050}} & \valmeta{\textbf{0.062}} & \valmeta{\textbf{0.062}} & \valmeta{\textbf{0.068}} & \valmeta{\textbf{0.069}} & \valmeta{\textbf{0.107}} & \valmeta{\textbf{0.095}} & \valmeta{\textbf{0.129}} & \valmeta{\textbf{0.102}} \\
Pair\_Accuracy↑       & \valmeta{\textbf{0.932}} & \valmeta{\textbf{0.958}} & \valmeta{\textbf{0.970}} & \valmeta{\textbf{0.975}} & \valmeta{\textbf{0.980}} & \valmeta{\textbf{0.981}} & \valmeta{\textbf{0.984}} & \valmeta{\textbf{0.985}} & \valmeta{\textbf{0.987}} & \valmeta{\textbf{0.987}} \\
NDCG@50↑        & \valmeta{\textbf{0.992}} & \valmeta{\textbf{0.996}} & \valmeta{0.996}           & \valmeta{\textbf{0.998}} & \valmeta{\textbf{0.999}} & \valmeta{\textbf{0.999}} & \valmeta{\textbf{0.999}} & \valmeta{\textbf{0.999}} & \valmeta{\textbf{0.999}} & \valmeta{\textbf{0.999}} \\
Top50\_Accuracy↑     & \valmeta{\textbf{0.960}} & \valmeta{0.920}           & \valmeta{0.940}           & \valmeta{\textbf{0.940}} & \valmeta{\textbf{0.940}} & \valmeta{\textbf{0.980}} & \valmeta{\textbf{0.960}} & \valmeta{\textbf{0.960}} & \valmeta{\textbf{0.960}} & \valmeta{\textbf{0.980}} \\
Ranking\_Error\_within 1\%↑      & \valmeta{\textbf{0.257}} & \valmeta{\textbf{0.337}} & \valmeta{\textbf{0.392}} & \valmeta{\textbf{0.492}} & \valmeta{\textbf{0.520}} & \valmeta{\textbf{0.552}} & \valmeta{\textbf{0.620}} & \valmeta{\textbf{0.645}} & \valmeta{\textbf{0.666}} & \valmeta{\textbf{0.677}} \\
Ranking\_Error\_within 2\%↑     & \valmeta{\textbf{0.432}} & \valmeta{\textbf{0.535}} & \valmeta{\textbf{0.645}} & \valmeta{\textbf{0.725}} & \valmeta{\textbf{0.788}} & \valmeta{\textbf{0.818}} & \valmeta{\textbf{0.865}} & \valmeta{\textbf{0.871}} & \valmeta{\textbf{0.910}} & \valmeta{\textbf{0.913}} \\
Ranking\_Error\_within 5\%↑     & \valmeta{\textbf{0.738}} & \valmeta{\textbf{0.869}} & \valmeta{\textbf{0.964}} & \valmeta{\textbf{0.974}} & \valmeta{\textbf{0.989}} & \valmeta{\textbf{0.988}} & \valmeta{\textbf{0.997}} & \valmeta{\textbf{1.000}} & \valmeta{\textbf{1.000}} & \valmeta{\textbf{1.000}} \\
Ranking\_Error\_within 10\%↑     & \valmeta{\textbf{0.946}} & \valmeta{\textbf{0.988}} & \valmeta{\textbf{0.998}} & \valmeta{\textbf{1.000}} & \valmeta{\textbf{1.000}} & \valmeta{\textbf{1.000}} & \valmeta{\textbf{1.000}} & \valmeta{\textbf{1.000}} & \valmeta{\textbf{1.000}} & \valmeta{\textbf{1.000}} \\
\bottomrule
\end{tabular}
}
\end{table*}

\begin{table*}[tb!]
\centering
\small
\setlength{\tabcolsep}{6pt} 
\renewcommand{\arraystretch}{1.2}
\caption{MetaBench vs. EssenceBench on ARC dataset \footnotesize(↑ larger is better, ↓ smaller is better)}
\label{tab:arc}
\resizebox{\textwidth}{!}{
\begin{tabular}{l|*{10}{r}}
\toprule
\multirow{2}{*}{\textbf{Method \& Metric}} & \multicolumn{10}{c}{\textbf{ARC Coreset Size}}\\
\cmidrule(l){2-11}&
\multicolumn{1}{c}{\;\;\;50}  & \multicolumn{1}{c}{\;\;\;100} & \multicolumn{1}{c}{\;\;\;150} &  \multicolumn{1}{c}{\;\;\;200} & \multicolumn{1}{c}{\;\;\;250} & \multicolumn{1}{c}{\;\;\;300} & \multicolumn{1}{c}{\;\;\;350} & \multicolumn{1}{c}{\;\;\;400} & \multicolumn{1}{c}{\;\;\;450} & \multicolumn{1}{c}{\;\;\;500} \\

\rowcolor{metaBG}
\multicolumn{11}{l}{\textbf{MetaBench}}\\
RMSE↓               & \valmeta{2.968} & \valmeta{2.082} & \valmeta{1.690} & \valmeta{1.511} & \valmeta{1.332} & \valmeta{1.186} & \valmeta{1.077} & \valmeta{0.961} & \valmeta{0.890} & \valmeta{0.836} \\
Pearson↑            & \valmeta{0.986} & \valmeta{0.993} & \valmeta{0.995} & \valmeta{0.996} & \valmeta{0.997} & \valmeta{0.998} & \valmeta{0.998} & \valmeta{0.998} & \valmeta{0.999} & \valmeta{0.999} \\
Spearman↑           & \valmeta{0.976} & \valmeta{0.984} & \valmeta{0.991} & \valmeta{0.992} & \valmeta{0.994} & \valmeta{0.995} & \valmeta{0.996} & \valmeta{0.997} & \valmeta{0.997} & \valmeta{0.998} \\
Kendall↑            & \valmeta{0.870} & \valmeta{0.898} & \valmeta{0.922} & \valmeta{0.928} & \valmeta{0.936} & \valmeta{0.942} & \valmeta{0.949} & \valmeta{0.955} & \valmeta{0.959} & \valmeta{0.961} \\
Rank\_Stability↑         & \valmeta{0.014} & \valmeta{0.021} & \valmeta{0.026} & \valmeta{0.027} & \valmeta{0.041} & \valmeta{0.035} & \valmeta{0.038} & \valmeta{0.050} & \valmeta{0.044} & \valmeta{0.044} \\
Pair\_Accuracy↑         & \valmeta{0.918} & \valmeta{0.940} & \valmeta{0.954} & \valmeta{0.959} & \valmeta{0.963} & \valmeta{0.967} & \valmeta{0.972} & \valmeta{0.974} & \valmeta{0.977} & \valmeta{0.978} \\
NDCG@50↑           & \valmeta{\textbf{0.987}} & \valmeta{0.992} & \valmeta{0.995} & \valmeta{0.994} & \valmeta{0.997} & \valmeta{0.998} & \valmeta{0.998} & \valmeta{\textbf{0.999}} & \valmeta{0.999} & \valmeta{0.998} \\
Top50\_Accuracy↑        & \valmeta{\textbf{0.900}} & \valmeta{0.880} & \valmeta{0.900} & \valmeta{0.900} & \valmeta{0.880} & \valmeta{0.880} & \valmeta{0.920} & \valmeta{0.920} & \valmeta{0.900} & \valmeta{0.960} \\
Ranking\_Error\_within 1\%↑         & \valmeta{0.150} & \valmeta{0.236} & \valmeta{0.274} & \valmeta{0.317} & \valmeta{0.368} & \valmeta{0.361} & \valmeta{0.412} & \valmeta{0.459} & \valmeta{0.456} & \valmeta{0.492} \\
Ranking\_Error\_within 2\%↑        & \valmeta{0.328} & \valmeta{0.442} & \valmeta{0.492} & \valmeta{0.552} & \valmeta{0.597} & \valmeta{0.615} & \valmeta{0.669} & \valmeta{0.704} & \valmeta{0.753} & \valmeta{0.767} \\
Ranking\_Error\_within 5\%↑        & \valmeta{0.653} & \valmeta{0.756} & \valmeta{0.844} & \valmeta{0.853} & \valmeta{0.887} & \valmeta{0.902} & \valmeta{0.952} & \valmeta{0.956} & \valmeta{0.974} & \valmeta{0.970} \\
Ranking\_Error\_within 10\%↑        & \valmeta{0.881} & \valmeta{0.929} & \valmeta{0.973} & \valmeta{0.985} & \valmeta{0.985} & \valmeta{0.992} & \valmeta{0.998} & \valmeta{1.000} & \valmeta{1.000} & \valmeta{0.998} \\

\midrule
\rowcolor{essBG}
\multicolumn{11}{l}{\textbf{EssenceBench}}\\
RMSE↓               & \valmeta{\textbf{2.045}} & \valmeta{\textbf{1.104}} & \valmeta{\textbf{0.841}} & \valmeta{\textbf{0.703}} & \valmeta{\textbf{0.612}} & \valmeta{\textbf{0.529}} & \valmeta{\textbf{0.501}} & \valmeta{\textbf{0.422}} & \valmeta{\textbf{0.368}} & \valmeta{\textbf{0.347}} \\
Pearson↑            & \valmeta{\textbf{0.993}} & \valmeta{\textbf{0.998}} & \valmeta{\textbf{0.999}} & \valmeta{\textbf{0.999}} & \valmeta{\textbf{0.999}} & \valmeta{\textbf{1.000}} & \valmeta{\textbf{1.000}} & \valmeta{\textbf{1.000}} & \valmeta{\textbf{1.000}} & \valmeta{\textbf{1.000}} \\
Spearman↑           & \valmeta{\textbf{0.988}} & \valmeta{\textbf{0.996}} & \valmeta{\textbf{0.997}} & \valmeta{\textbf{0.998}} & \valmeta{\textbf{0.998}} & \valmeta{\textbf{0.999}} & \valmeta{\textbf{0.999}} & \valmeta{\textbf{0.999}} & \valmeta{\textbf{0.999}} & \valmeta{\textbf{0.999}} \\
Kendall↑            & \valmeta{\textbf{0.907}} & \valmeta{\textbf{0.947}} & \valmeta{\textbf{0.958}} & \valmeta{\textbf{0.964}} & \valmeta{\textbf{0.969}} & \valmeta{\textbf{0.973}} & \valmeta{\textbf{0.974}} & \valmeta{\textbf{0.978}} & \valmeta{\textbf{0.981}} & \valmeta{\textbf{0.983}} \\
Rank\_Stability↑        & \valmeta{\textbf{0.035}} & \valmeta{\textbf{0.047}} & \valmeta{\textbf{0.063}} & \valmeta{\textbf{0.071}} & \valmeta{\textbf{0.077}} & \valmeta{\textbf{0.096}} & \valmeta{\textbf{0.084}} & \valmeta{\textbf{0.120}} & \valmeta{\textbf{0.147}} & \valmeta{\textbf{0.149}} \\
Pair\_Accuracy↑         & \valmeta{\textbf{0.939}} & \valmeta{\textbf{0.965}} & \valmeta{\textbf{0.973}} & \valmeta{\textbf{0.977}} & \valmeta{\textbf{0.980}} & \valmeta{\textbf{0.983}} & \valmeta{\textbf{0.983}} & \valmeta{\textbf{0.986}} & \valmeta{\textbf{0.988}} & \valmeta{\textbf{0.989}} \\
NDCG@50↑           & \valmeta{0.986} & \valmeta{\textbf{0.995}} & \valmeta{\textbf{0.999}} & \valmeta{\textbf{0.999}} & \valmeta{\textbf{0.999}} & \valmeta{\textbf{0.999}} & \valmeta{\textbf{1.000}} & \valmeta{\textbf{0.999}} & \valmeta{\textbf{1.000}} & \valmeta{\textbf{0.999}} \\
Top50\_Accuracy↑        & \valmeta{\textbf{0.900}} & \valmeta{\textbf{0.940}} & \valmeta{\textbf{0.960}} & \valmeta{\textbf{0.920}} & \valmeta{\textbf{0.960}} & \valmeta{\textbf{1.000}} & \valmeta{\textbf{0.960}} & \valmeta{\textbf{0.980}} & \valmeta{\textbf{0.960}} & \valmeta{0.940} \\
Ranking\_Error\_within 1\%↑         & \valmeta{\textbf{0.233}} & \valmeta{\textbf{0.408}} & \valmeta{\textbf{0.444}} & \valmeta{\textbf{0.498}} & \valmeta{\textbf{0.550}} & \valmeta{\textbf{0.630}} & \valmeta{\textbf{0.639}} & \valmeta{\textbf{0.696}} & \valmeta{\textbf{0.737}} & \valmeta{\textbf{0.771}} \\
Ranking\_Error\_within 2\%↑         & \valmeta{\textbf{0.433}} & \valmeta{\textbf{0.641}} & \valmeta{\textbf{0.735}} & \valmeta{\textbf{0.768}} & \valmeta{\textbf{0.823}} & \valmeta{\textbf{0.869}} & \valmeta{\textbf{0.871}} & \valmeta{\textbf{0.901}} & \valmeta{\textbf{0.937}} & \valmeta{\textbf{0.955}} \\
Ranking\_Error\_within 5\%↑         & \valmeta{\textbf{0.788}} & \valmeta{\textbf{0.928}} & \valmeta{\textbf{0.965}} & \valmeta{\textbf{0.986}} & \valmeta{\textbf{0.991}} & \valmeta{\textbf{0.997}} & \valmeta{\textbf{0.997}} & \valmeta{\textbf{0.998}} & \valmeta{\textbf{1.000}} & \valmeta{\textbf{1.000}} \\
Ranking\_Error\_within 10\%↑        & \valmeta{\textbf{0.949}} & \valmeta{\textbf{0.997}} & \valmeta{\textbf{1.000}} & \valmeta{\textbf{1.000}} & \valmeta{\textbf{1.000}} & \valmeta{\textbf{1.000}} & \valmeta{\textbf{1.000}} & \valmeta{\textbf{1.000}} & \valmeta{\textbf{1.000}} & \valmeta{\textbf{1.000}} \\
\bottomrule
\end{tabular}
}
\end{table*}

\section{Details of Experiments}
\subsection{Baseline Methods for Data Selection}
\label{appendix:baseline_method}
For baseline comparisons, we include the following widely used methods. Random Selection selects subsets by randomly sampling from the original data. GraNd selects the top-k data points based on the gradient norms of the final token in the prediction task, ranking samples in descending order of gradient magnitude. PPL (Perplexity) is a standard metric for evaluating language models, defined as the exponential of the average negative log-likelihood over the predicted tokens, and is computed over the full predicted sequence for each question in the dataset. To compute the gradients for GraNd-based selection, we use the Llama-3.1-8B-Instruct\footnote{https://huggingface.co/meta-llama/Llama-3.1-8B-Instruct} model as the scoring backbone.

\subsection{Details of Dataset Construction}
\label{appendix:dataset_construction}
Following the initial collection of benchmark data, we performed a coarse filtering step to remove redundant or uninformative items before subset selection. This stage was guided by two primary criteria: textual similarity and ranking consistency.

To mitigate redundancy in the benchmark, we applied coarse filtering using both semantic and behavioral criteria. For semantic overlap, we used the \texttt{bge-m3} model~\citep{bge} to embed each item and computed pairwise similarities across all examples. Items with high embedding similarity were removed to ensure lexical and conceptual diversity. In parallel, we identified items with redundant behavioral signals by examining their ranking patterns over LLMs. Items that yielded highly correlated rankings were pruned, as they offered limited additional insight into model differences. Together, these two filters reduced both surface-level and functional redundancy, yielding a more informative candidate pool for downstream compression.

In the case of the \texttt{MMLU} dataset, we observed that many items exhibited naturally high textual and behavioral similarity due to the curriculum-style structure of the benchmark. To avoid over-pruning, we adapted the filtering thresholds to be more lenient for MMLU.

\end{document}